\documentclass{acmart} 
\usepackage{array}
\usepackage{rotating}
\usepackage{graphicx}
\usepackage{amsmath}
\usepackage{multirow}
\usepackage{multicol}
\usepackage{longtable}
\usepackage{siunitx}
\usepackage{booktabs}
\usepackage{adjustbox}
\usepackage{pifont}
\usepackage{tabularx}
\usepackage{overpic}
\usepackage{tabularx}
\usepackage{multirow}
\usepackage{placeins}
\usepackage{verbatim} 
\usepackage{algpseudocode}
\usepackage{cases}
\usepackage{ltablex}
\usepackage{algorithm}
\usepackage{subcaption}
\usepackage{geometry}
\usepackage{wrapfig}
\usepackage{tikz}
\usetikzlibrary{positioning,arrows.meta}

\newcolumntype{L}{>{\RaggedRight\hspace{0pt}}X} 
\newcommand{\etal}{\textit{et al.}}

\AtBeginDocument{%
  }

\setcopyright{acmcopyright}
\copyrightyear{2018}
\acmYear{2018}
\acmDOI{XXXXXXX.XXXXXXX}

\acmJournal{JACM}
\acmVolume{37}
\acmNumber{4}
\acmArticle{111}
\acmMonth{8}

\begin{document}

\title{Quantization in Federated Learning: Methods, Challenges and Future Directions}

\author{Farwa Ikram}
\email{farwa.ikram@dimes.unical.it}
\orcid{XXXX-XXXX-XXXX-XXXX}
\author{Dipanwita Thakur}
\email{dipanwita.thakur@unical.it}
\orcid{0000-0003-2895-1425}
\author{Antonella Guzzo}
\email{antonella.guzzo@unical.it}
\author{Giancarlo Fortino}
\email{giancarlo.fortino@unical.it}
\affiliation{%
  \institution{DIMES - University of Calabria}
  \streetaddress{Via P. Bucci, cubo 41c}
  \city{Rende}
  \state{Cosenza (CS)}
  \country{Italy}
  \postcode{87036}
}

\renewcommand{\shortauthors}{Ikram et al.}

\begin{abstract}
Federated Learning (FL) has become a foundational paradigm for privacy-preserving distributed intelligence, yet its scalability remains fundamentally constrained by communication bottlenecks, device heterogeneity, and the challenges of training under statistically non-IID data. Quantization is one of the most effective mechanisms for mitigating these limitations, reducing both uplink/downlink payloads and on-device computation. This paper provides the first FL-centric systematic review of quantization, introducing a novel taxonomy organized around FL-specific dimensions, including client heterogeneity, aggregation consistency, communication-scheduling adaptation, non-IID robustness, privacy/security integration, and hardware/energy co-optimization. Beyond cataloging existing methods, we analyze how quantization interacts with core FL behaviors such as client drift, partial participation, convergence stability, secure aggregation, and differential privacy. We further identify cross-method insights, open research gaps, and design guidelines for practitioners deploying quantized FL on mobile, IoT, and edge platforms. This survey thus establishes quantization not merely as a compression technique, but as a fundamental systems component shaping the performance, robustness, and practicality of modern FL.
\end{abstract}



\received{20 February 2007}
\received[revised]{12 March 2009}
\received[accepted]{5 June 2009}

\maketitle

\section{Introduction}
The rapid growth of Internet of Things (IoT) devices, mobile systems, and embedded sensors has enabled large-scale distributed intelligence at the network edge. While these devices generate vast amounts of data, transmitting raw data to centralized servers for model training raises serious concerns related to privacy, bandwidth, latency, and energy consumption. Federated Learning (FL)~\cite{McMahan2017} addresses these challenges by enabling multiple clients to collaboratively train a shared global model without sharing local data. By keeping data localized, FL reduces communication overhead and preserves privacy, making it well suited for edge and mobile environments. 
Despite its advantages, deploying FL in real-world settings remains challenging due to device heterogeneity, limited computational resources, and high communication costs. In large-scale FL systems, the exchange of model updates between clients and the server often dominates training time and energy consumption, leading to latency, unstable convergence, and reduced participation of resource-constrained devices~\cite{Zhao2022}. Consequently, improving communication efficiency is a central challenge in scalable FL.

To address this issue, numerous communication-efficient techniques have been proposed, including client scheduling, sparsification, pruning, knowledge distillation, low-rank approximation, and quantization~\cite{Konečný2017, Reisizadeh2020, Thakur2025}. However, many of these methods introduce additional overhead or complexity. For example, pruning requires transmitting masks, knowledge distillation demands additional training pipelines~\cite{Wu2022}, and sparsification may degrade convergence under non-IID data. In contrast, quantization consistently reduces both communication and computation cost by lowering numerical precision, making it particularly attractive for resource-constrained and energy-limited FL deployments.

Model quantization compresses model parameters, gradients, or activations into low-bit representations, enabling efficient communication and computation on edge devices. While quantization improves efficiency, it also introduces challenges such as accuracy degradation, convergence instability, and sensitivity to data heterogeneity~\cite{Khan2021}. Understanding these trade-offs is therefore critical for designing robust and scalable FL systems.

\subsection{Comparison with Related Survey and Motivation}
A large number of surveys have studied FL from diverse perspectives, including FL fundamentals~\cite{Aledhari2020, Li2020, Liu2020, Liu2024, Lo2021, Ren2025, Wahab2021, Yang2019, Zhang2021}, mobile and edge FL~\cite{Duan2023, Jia2025, Lim2020, Wu2024}, communication efficiency~\cite{Almanifi2023, Liu2020, Niknam2020, Shahid2021, Zhao2023}, heterogeneity~\cite{Pfeiffer2023, Ye2023}, privacy and security~\cite{Chen2023, Khraisat2024, Kumar2024, Shao2024}, and green FL~\cite{Thakur2025}. As summarized in Table~\ref{tab:surveyFLchallenges}, it is evident that existing surveys on FL primarily concentrate on isolated challenges such as privacy and security, statistical and system heterogeneity, communication efficiency, computation efficiency, and energy efficiency. These surveys typically treat quantization as a subcomponent of communication efficiency or model compression. However, no existing survey provides a dedicated and systematic review of quantization techniques in FL. Prior works do not analyze quantization as a standalone design dimension, nor do they provide a comprehensive taxonomy of quantization strategies and their trade-offs in terms of accuracy, convergence, communication cost, and system heterogeneity. In contrast, this paper is, to the best of our knowledge, the first survey exclusively focused on quantization in FL, offering a unified and comparative synthesis across algorithmic, system, and security dimensions.

\begin{table*}[t]
\centering
\footnotesize
\setlength{\tabcolsep}{4pt}
\renewcommand{\arraystretch}{1.15}
\caption{Comparison of Existing Surveys Related to Federated Learning}
\label{tab:surveyFLchallenges}
\begin{tabular}{|p{0.18\textwidth}|p{0.20\textwidth}|p{0.28\textwidth}|p{0.26\textwidth}|}
\hline
\textbf{References} & \textbf{Focus Area} & \textbf{Key Contributions} & \textbf{Limitations} \\
\hline

\cite{Aledhari2020, Li2020, Liu2022, Liu2024, Lo2021, Ren2025, Wahab2021, Yang2019, Zhang2021} 
& FL Fundamentals 
& Introduce FL concepts, architectures, and applications. 
& Do not analyze communication efficiency or quantization. \\
\hline

\cite{Duan2023, Jia2025, Lim2020, Liu2020, Niknam2020, Wu2024} 
& FL in Wireless / Edge Networks 
& Study FL integration with edge, 5G/6G, and wireless systems. 
& Limited discussion of quantized model compression. \\
\hline

\cite{Khan2021,Nguyen2021} 
& FL for IoT 
& Survey FL-enabled IoT services and applications. 
& Lack in-depth analysis of communication and quantization. \\
\hline

\cite{Qi2024, Sah2022} 
& Model Aggregation 
& Review aggregation strategies and convergence issues. 
& Communication compression and energy efficiency not addressed. \\
\hline

\cite{Almanifi2023, Liu2020, Niknam2020, Shahid2021, Zhao2023} 
& Communication-Efficient FL 
& Review sparsification, quantization, scheduling, and client selection. 
& Quantization analysis remains high-level. \\
\hline

\cite{Pfeiffer2023, Ye2023} 
& Heterogeneity-Aware FL 
& Address statistical, device, and system heterogeneity. 
& Energy efficiency and adaptive quantization underexplored. \\
\hline

\cite{Chen2023,Shao2024,Kumar2024,Khraisat2024} 
& Privacy and Security in FL 
& Analyze privacy risks, attacks, and secure aggregation. 
& Limited treatment of quantization–privacy interactions. \\
\hline

\cite{Thakur2025} 
& Green FL 
& Study energy efficiency and carbon-aware FL. 
& Quantization techniques not considered. \\
\hline

\textbf{Ours} 
& Quantization in FL 
& Systematic survey and taxonomy of quantization techniques in FL. 
& Addresses communication, robustness, heterogeneity, and privacy. \\
\hline

\end{tabular}
\end{table*}

As evidenced by the growing number of recent studies on quantized FL, there is a clear need for a systematic review that consolidates existing knowledge, identifies limitations, and outlines future research directions. This motivates the present PRISMA-based systematic literature review. Furthermore, the number of publications related to quantization in FL for improving effectiveness and efficiency has increased considerably in recent years. This growing body of work highlights the need for a systematic review that synthesizes existing knowledge, analyzes methodological strengths and limitations, and outlines future research opportunities. Therefore, this paper proposes a systematic literature review (SLR) focused on examining quantization techniques within the scope of enhancing effectiveness and efficiency in FL.

This systematic review is designed to address the following research questions (RQs): \textbf{RQ1:} How do different quantization methods impact the performance, efficiency, and effectiveness of FL systems? \textbf{RQ2:} What challenges and limitations characterize existing quantization approaches in FL? \textbf{RQ3:} What research gaps persist across algorithmic, statistical, and system dimensions of quantized FL? and \textbf{RQ4:} What emerging trends can guide the design of scalable, robust, and privacy-aware quantized FL systems? By systematically addressing these research questions, this survey provides the first systematic synthesis of quantization methods in FL, offering a new perspective that bridges communication efficiency, computation cost, and model performance. This work thereby lays a foundation for developing scalable, communication-efficient, and resource-aware FL systems in future research and industrial applications.
\subsection{Contributions}
In this survey, we present a PRISMA-based systematic review of model quantization in FL. Specifically, we synthesize and critically analyze existing research to evaluate the impact, challenges, gaps, and future directions of quantization techniques within FL frameworks. This work gives a novel taxonomy and serves as a structured guide for researchers and practitioners aiming to understand and apply quantization methods to improve communication efficiency, model performance, and scalability in FL. To the best of our knowledge, this article constitutes the first systematic and holistic review focusing exclusively on model quantization in FL. The key contributions are summarized as follows:

\begin{itemize}

  \item \textbf{Comprehensive synthesis of quantization techniques in FL (RQ1):}  
    We systematically categorize and analyze quantization methods from an FL-centric perspective to provide design dimensions and evaluate their impact on six different system-level properties, such as system heterogeneity, communication efficiency, non-iid robustness, aggregation consistency, privacy/security integration, and hardware-energy co-optimization to introduce a novel taxonomy.
  \item \textbf{Critical evaluation of limitations and trade-offs (RQ2):}  
  We identify and assess the primary challenges faced by existing quantized FL approaches, including communication overhead, system heterogeneity, scalability, aggregation consistency, and convergence instability under non-IID data, and privacy–utility trade-offs. 
  \item \textbf{Identification of research gaps and methodological deficiencies (RQ3):}  
  Through systematic mapping along FL-specific dimensions, we uncover critical gaps in the literature, including the lack of standardized benchmarks, limited evaluation under realistic non-IID and heterogeneous settings, and insufficient exploration of adaptive quantization strategies co-designed with aggregation, scheduling, and system constraints.
   
  \item \textbf{Exploration of emerging trends and future research directions (RQ4):}  
  We highlight promising directions for advancing quantized FL, including adaptive precision mechanisms, joint design with aggregation and scheduling policies, integration with privacy-preserving techniques, and hardware- and energy-aware optimization for mobile, IoT, and edge platforms.

\end{itemize}
\begin{figure*}
    \centering
    \includegraphics[width=\linewidth,height=5.5in,keepaspectratio=FALSE]{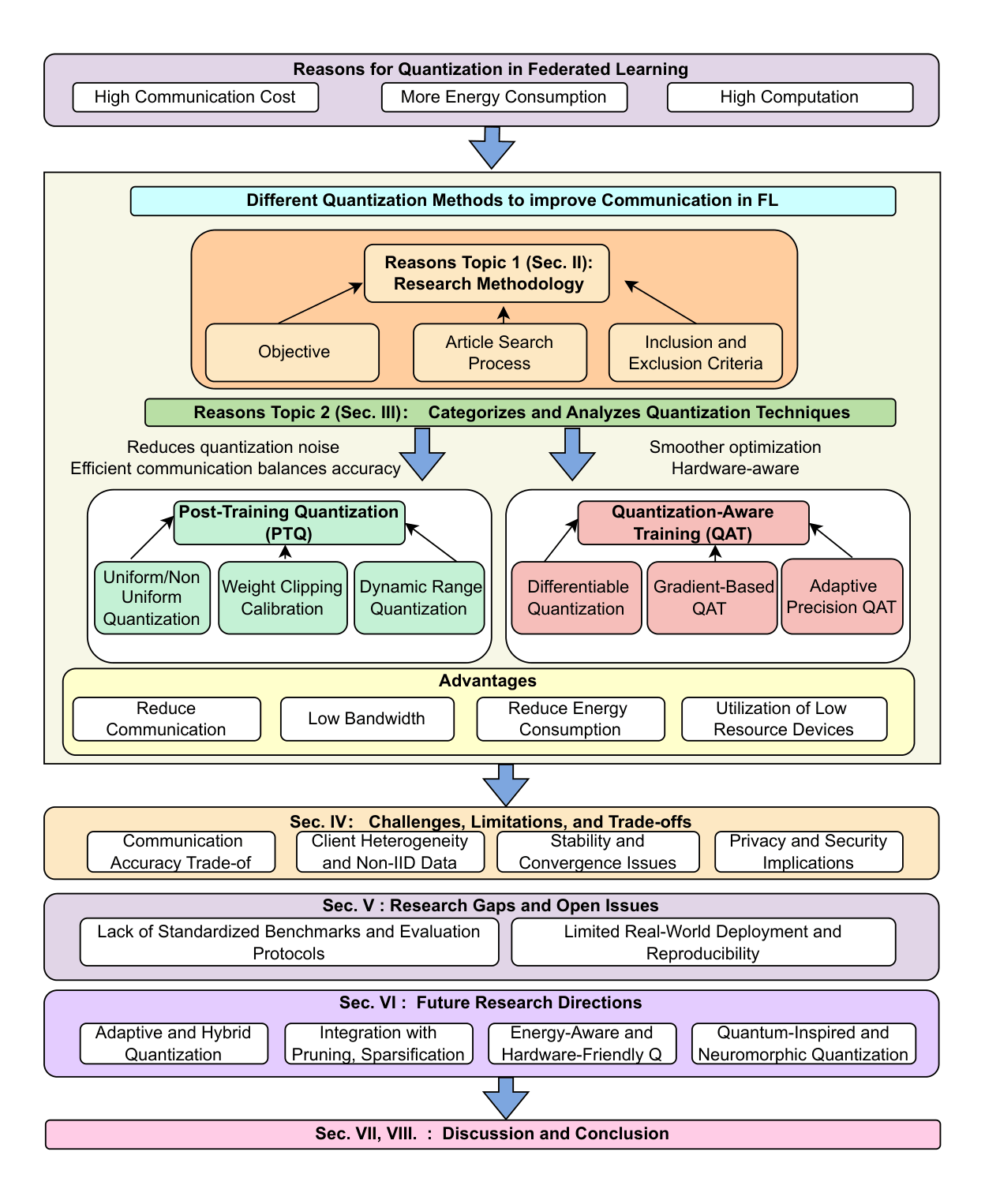}
    \vspace{-0.8cm}
    \caption{Structure of the Survey}
    \label{fig:structure}
    \vspace{-0.5cm}
\end{figure*}
\vspace{-0.3cm}
\subsection{Structure of the Paper}
The scope and organization of the survey is shown in Figure \ref{fig:structure}. Section~\ref{sec:methodology} outlines the PRISMA-based review methodology. Section~\ref{sec:quantization} categorizes and analyzes quantization techniques in FL, addressing \textbf{RQ1}. Section~\ref{sec:challenges} discusses existing challenges and trade-offs (\textbf{RQ2}), while Section~\ref{sec:gaps} highlights key research gaps (\textbf{RQ3}). Section~\ref{sec:future} presents promising future research directions (\textbf{RQ4}). Finally, Sections~\ref{sec:discussion} and~\ref{sec:conclusion} provide an integrated discussion and concluding remarks.

\section{Research Methodology}
\label{sec:methodology}
\subsection{Objective}
The proposed work is a systematic literature review (SLR) using PRISMA (Preferred Reporting Items for Systematic Reviews)~\cite{Moher2010}. Figure \ref{fig:fig2} shows a complete flowchart of the proposed methodology. In the first step of SLR, research questions are defined, and then relevant articles are extracted using appropriate search strings. Selected articles are then filtered based on inclusion-exclusion criteria to remove irrelevant studies, and then the remaining articles are analyzed in detail.

\begin{table*}[htbp]
\centering
\caption{Number of studies retrieved from each database}
\resizebox{\textwidth}{!}{
\begin{tabular}{p{3.2cm} p{5.7cm} p{3.2cm} p{3.2cm}}
\hline
        Database & Search String & Initial number of studies retrieved & Eligible studies after screening \\ \hline
        IEEE Xplore & "All Metadata":Federated learning) AND ("All Metadata":Quantization) AND ("All Metadata":Communication)& 647 & 41 \\ \hline
        arXiv & ALL=(federated learning AND quantization AND communication) & 18 & 6 \\ \hline
        Science Direct & Federated learning AND quantization AND communication & 1593 & 2 \\ \hline
        ACM Digital Library & Federated learning AND Quantization & 1135 & 2 \\ \hline
        Wiley Online Library & All: federated learning] AND [All: quantization] AND [All: communication] & 454 & 1 \\ \hline
        Others & ``Federated learning'' and ``quantization'' and ``Communication'' & 78 & 8\\
        \hline
\end{tabular}
}
\label{tab:table2}
\vspace{-0.5cm}
\end{table*}

\subsection{Article Search Process}
\begin{figure}[t]
  \centering
  \begin{subfigure}{.3\columnwidth}
    \centering
    \includegraphics[width=\linewidth]{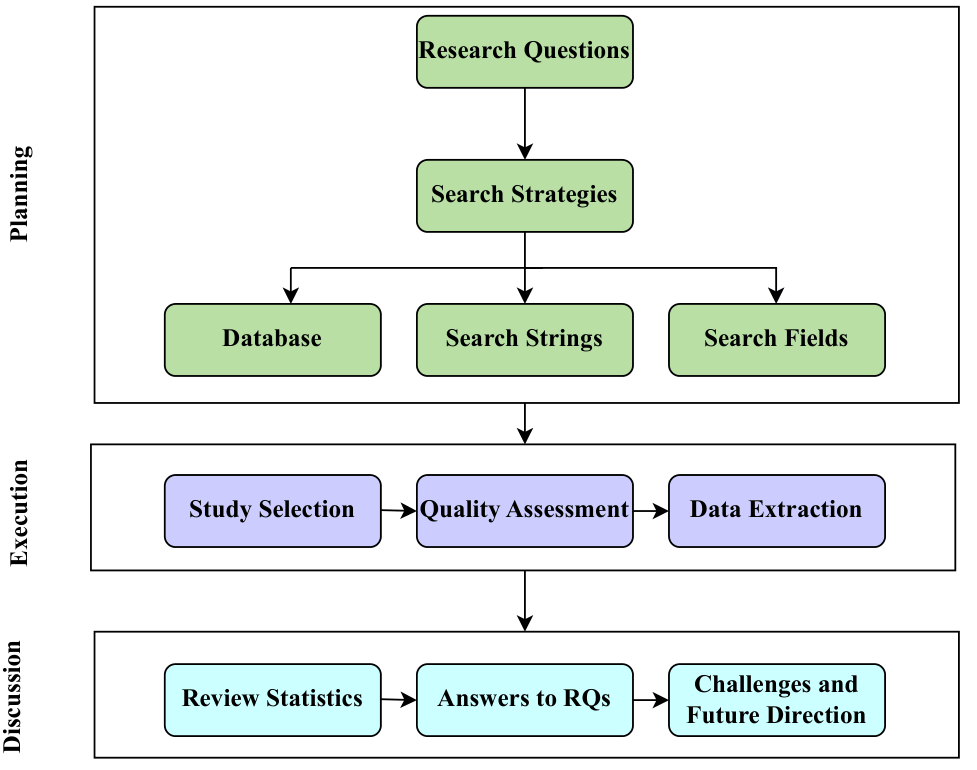}
    \caption{Flowchart of Proposed SLR Framework}
    \label{fig:fig2}
  \end{subfigure}%
  \hfill
  \begin{subfigure}{.3\columnwidth}
    \centering
    \includegraphics[width=\linewidth]{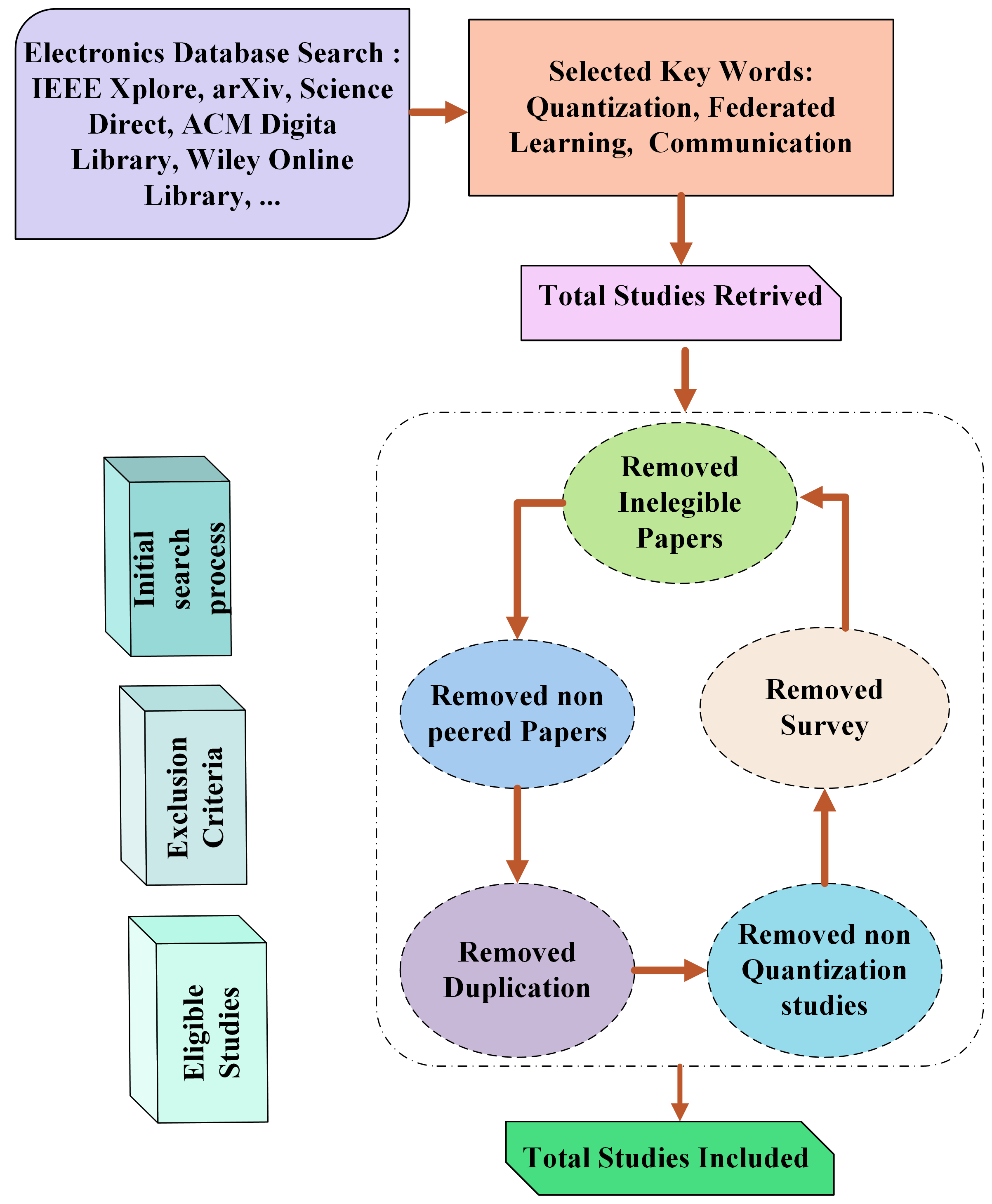}
    \caption{Proposed SLR Protocol}
\label{fig:fig3}
  \end{subfigure}%
  \hfill
  \begin{subfigure}{.3\columnwidth}
    \centering
    \includegraphics[width=\linewidth]{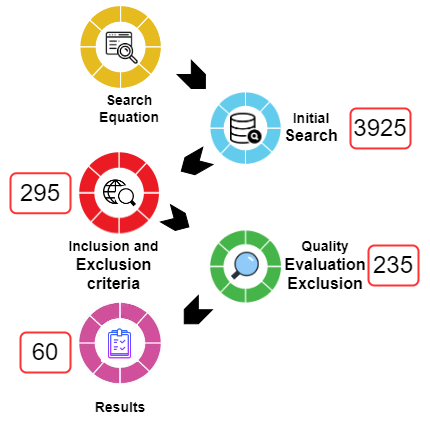}
    \caption{Article Extraction Process}
\label{fig:fig4}
  \end{subfigure}
  \caption{}
  \vspace{-0.5cm}
\end{figure}

 We consider standard databases with a custom range of year 2019-2026. Advanced options for different databases are also used to extract all the relevant articles. The search strings, we used are depicted in Table \ref{tab:table2} along with an initial number of studies retrieved in each database. The search starts with the title, abstract, and keywords in each of the mentioned databases. Abstracts are analyzed, and irrelevant articles are discarded at this stage. The overview of the search process to extract articles is depicted in Figure \ref{fig:fig3}.

\subsection{Inclusion and Exclusion Criteria}
The search process results in the extraction of articles, including irrelevant articles. To remove the irrelevant articles, an inclusion and exclusion criterion is defined. This procedure filters the articles to only those that include specific types of studies relevant to our problem. The inclusion and exclusion criteria are defined to ensure relevance and quality of the selected studies. Specifically, we include peer-reviewed articles written in English that investigate quantization as a primary research focus within FL frameworks, with particular attention to its role in improving communication-computation and energy efficiency or system performance. Included articles are quantization-related publications in flagship conferences or reputable journals. Figure \ref{fig:fig4} shows the number of articles extracted at each stage of the process. 

\section{Quantization in FL: State of the Art}
\label{sec:quantization}

\begin{figure}
    \centering
    \includegraphics[width=0.9\textwidth]{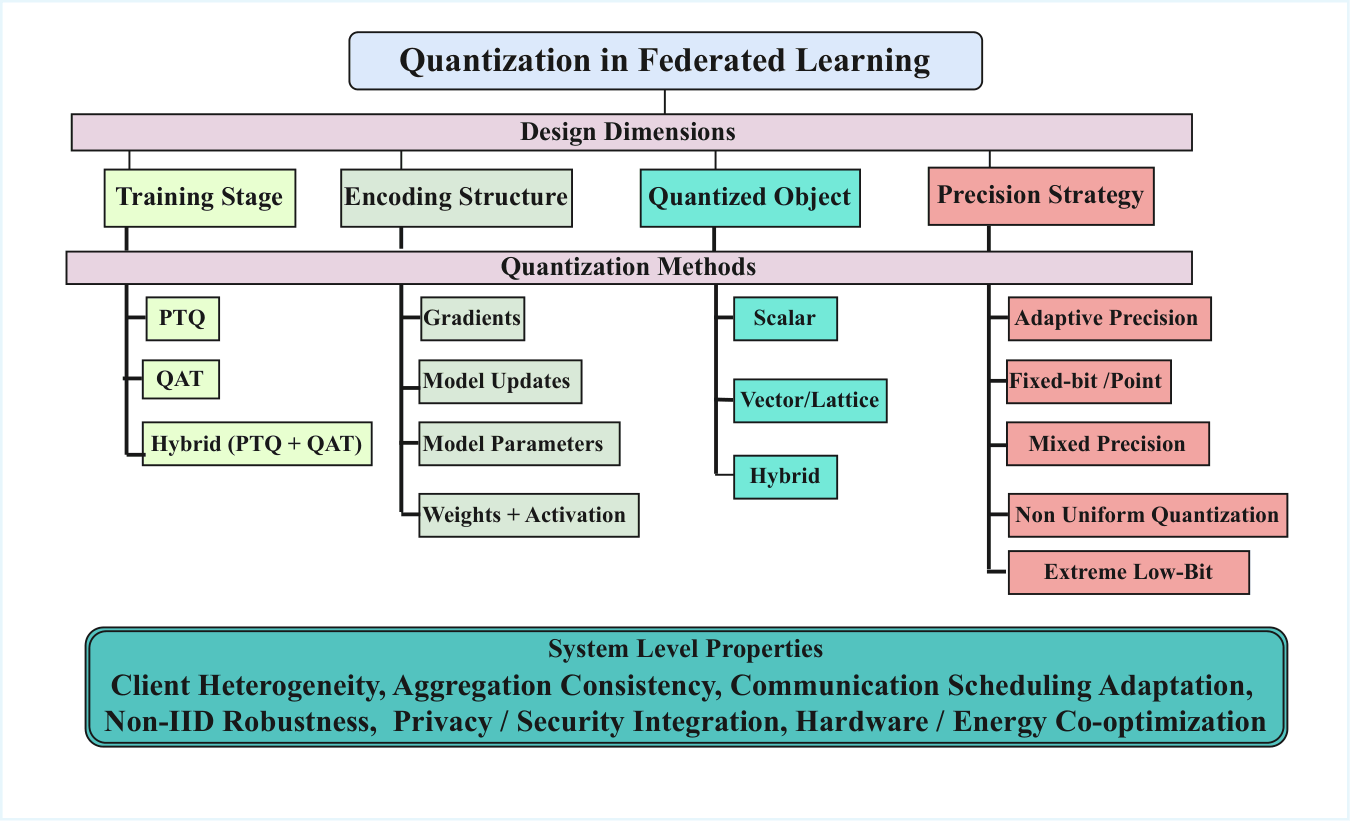}
    \vspace{-0.3cm}
    \caption{Design Dimensions for Quantization Taxonomy in FL}
    \label{fig:designdimension}
    \vspace{-0.5cm}
\end{figure}

Quantization plays a pivotal role in FL by reducing the communication overhead associated with transmitting model updates between clients and the central server. In large-scale FL systems, the size of model parameters or gradients can be prohibitively large, leading to communication bottlenecks, energy inefficiency, and slow convergence. Quantization mitigates these challenges while maintaining model accuracy and convergence stability~\cite{Alistarh2017uniform, Reisizadeh2020, Shlezinger2020_a}. In this section, we discuss quantization methods that are not intended to be mutually exclusive. Instead, they represent complementary design dimensions along which quantization can be applied in FL systems. Specifically, these dimensions include (i) the training stage at which quantization is introduced (post-training quantization versus quantization-aware training), (ii) the object being quantized (model parameters, gradients, or model updates), (iii) the encoding structure (scalar versus vector, uniform vs non-uniform), and (iv) the precision strategy (fixed-bit, variable-bit, or adaptive precision) as shown in Figure \ref{fig:designdimension}. Each dimension corresponds to a distinct mechanism that affects optimization behavior, error propagation, communication efficiency, and convergence in different ways. Consequently, methods drawn from different dimensions may be combined within a single FL algorithm, while still constituting fundamentally different quantization approaches. These dimensions capture where quantization is applied in the learning pipeline (training stage), what is quantized (object), how values are encoded (encoding structure), and with what precision control strategy (precision). Together, they define the fundamental algorithmic choices underlying quantized FL methods and serve as the basis for the taxonomy presented in this work to address \textbf{RQ1}.

In addition, the diagram highlights system-level properties—including client heterogeneity (CH), aggregation consistency (AC), communication scheduling adaptation (CSA), robustness to non-IID data (NIR), privacy and security integration (PI), and hardware/energy co-optimization (HE), that influence the practical deployment of quantized FL systems.

\paragraph{Post-Training Quantization (PTQ)}
PTQ is one of the most practical and widely adopted model compression techniques for deploying deep neural networks on resource-constrained edge devices. PTQ applies quantization after model training, eliminating the need for retraining or fine-tuning~\cite{Gholami2021, Dubey2024uniform}. PTQ is computationally efficient but may cause minor accuracy loss, especially under non-IID data~\cite{krishnamoorthi2018quantizing, bozorgasl2025communication}. It converts full-precision (e.g., 32-bit floating-point) parameters and activations into low-bit representations (e.g., 8-bit integers) to reduce model size, memory footprint, and inference latency~\cite{Gholami2021, krishnamoorthi2018quantizing}. PTQ techniques rely on calibration or statistical approximations to estimate quantization parameters (scale and zero-point) that best represent the original data distribution. The main advantage of PTQ is its simplicity and compatibility with existing pretrained models, making it ideal for large-scale deployment and FL environments where retraining is costly or impractical. However, PTQ typically incurs a slight accuracy drop compared to QAT, especially for very low bit-widths or non-uniform data distributions. In the context of \textit{FL}, PTQ provides a lightweight yet effective approach for communication-efficient model sharing. Clients can quantize their local model updates before transmitting them to the central server, thereby reducing communication bandwidth and energy consumption~\cite{Dubey2024uniform, bozorgasl2025communication}. 

\paragraph{Dynamic Range Quantization:} Dynamic Range Quantization (DRQ) is a post-training uniform scalar quantization scheme in which quantization scales are determined from the observed dynamic range (min–max statistics) of model parameters, while activations remain in floating-point precision. Unlike full integer quantization, where both weights and activations are quantized, DRQ performs quantization only on static model parameters and dynamically rescales activations during inference. This approach provides a favorable balance between computational efficiency and model accuracy~\cite{krishnamoorthi2018quantizing, Jacob2018CVPR}. The quantization process involves computing the dynamic range of each tensor, using the observed minimum and maximum values, and mapping them linearly into a lower precision format. The quantized integer values are dequantized during computation to maintain compatibility with mixed-precision hardware. Mathematically, given a weight tensor $W$, the quantization function can be expressed as $W_q = \mathrm{round}\!\left(\frac{W}{s}\right), \quad s = \frac{\max(W) - \min(W)}{2^{b} - 1},$ where $b$ denotes the bit-width (e.g., $b=8$), and $s$ is the scaling factor or quantization step size. The reconstructed tensor is obtained as $\hat{W} = s \times W_q$, which approximates the original full-precision weights. By quantizing only the weights and maintaining floating-point activations, DRQ minimizes model size and computation costs while avoiding accuracy degradation associated with fully quantized arithmetic. It provides up to $4\times$ reduction in model storage and $2$-$3\times$ faster inference with negligible accuracy loss~\cite{Gholami2021}. Moreover, DRQ requires no retraining, making it particularly effective for rapid model deployment.

In the context of \textit{FL}, DRQ is especially advantageous because it reduces the communication cost between clients and the central server. Quantizing weights before transmission reduces bandwidth usage and accelerates model aggregation without compromising accuracy~\cite{Dubey2024uniform, bozorgasl2025communication}. Since activations remain in floating-point form, local computation remains stable, which is crucial for heterogeneous devices with varying hardware capabilities. Modern frameworks such as TensorFlow Lite and PyTorch Mobile adopt dynamic range quantization as a default quantization method for edge deployment due to its simplicity and hardware compatibility~\cite{pytorch2023quantization}. Overall, DRQ offers a practical balance between model efficiency, communication reduction, and performance preservation. However, when activation distributions differ significantly across federated clients, DRQ may exhibit minor accuracy loss, suggesting the need for adaptive quantization strategies in future FL implementations.

\begin{table*}[htbp]
\centering
\caption{Comparison of Dynamic Range-based PTQ Schemes in FL and Neural Network Compression}

\resizebox{\textwidth}{!}{
\begin{tabular}{p{3.3cm} p{5.1cm} p{4.2cm} p{4.2cm} p{3.2cm}}
\hline
\textbf{Method/Contribution Area} & \textbf{Description/Approach} & \textbf{Advantages} & \textbf{Limitations} & \textbf{References} \\ 
\hline

\textbf{Standard DRQ} &
Quantizes model weights to INT8 while retaining activations in floating-point (FP16 or FP32). Scaling factors are computed from weight statistics, and activations are rescaled dynamically during inference. &  Reduces model size and memory footprint by up to 4$\times$; minimal accuracy degradation compared to full precision; compatible with mixed-precision hardware accelerators (e.g., CPUs, GPUs, TPUs).
 & Does not quantize activations, leaving potential for further compression; limited improvement in computational latency due to floating-point activations.  & \cite{krishnamoorthi2018quantizing, Jacob2018CVPR} \\

\hline

\textbf{DRQ in Edge/Mobile Frameworks} &
Adopted in deployment frameworks such as TensorFlow Lite and PyTorch Mobile, which apply post-training DRQ for efficient edge inference. Quantizes weights only and dequantizes on-the-fly. & Provides significant reduction in storage and inference cost for mobile/IoT devices;
supported across diverse hardware with minimal calibration.  &  Accuracy depends on representativeness of calibration data;
 no per-layer adaptation to different activation distributions.
 & \cite{pytorch2023quantization} \\

\hline

\textbf{DRQ for FL} &
Applies DRQ to compress model updates exchanged between clients and the server in FL. Weights are quantized to INT8 before transmission, while clients compute locally in FP32. & Reduces communication cost significantly in FL environments; maintains model accuracy across heterogeneous devices; requires no retraining, suitable for on-device FL. &
 Minor performance drop in highly non-IID data due to client-specific scaling; lacks adaptive scaling for client distribution shifts.
& \cite{Dubey2024uniform, bozorgasl2025communication} \\

\hline

\textbf{Hardware-Aware DRQ (Mixed-Precision Inference)} &
Integrates DRQ with adaptive mixed-precision inference, where layers sensitive to quantization retain higher precision (FP16) and robust layers use INT8 or lower. & Balances energy efficiency and accuracy; tailors quantization levels to layer sensitivity and hardware capability.  & Requires profiling to identify sensitivity of layers; complex to implement in distributed FL systems.
& \cite{Gholami2021, nvidia2021tensorrt} \\

\hline

\textbf{Adaptive Dynamic Range Quantization (ADRQ)} &
Extends DRQ by recalibrating scaling factors dynamically based on activation statistics or client data in FL rounds. Employs on-device adaptation to mitigate accuracy loss. & Improves generalization under client heterogeneity; reduces quantization bias across federated updates. & Requires additional computation per client for recalibration; not yet widely supported in mainstream FL frameworks.
 & \cite{krishnamoorthi2018quantizing, Dubey2024uniform} \\

\hline

\end{tabular}
}
\label{tab:dynamic_range_quantization_comparison}
\vspace{-0.5cm}
\end{table*}
Table \ref{tab:dynamic_range_quantization_comparison} summarizes key approaches to DRQ used in neural network compression and FL. While standard DRQ achieves memory and latency efficiency by quantizing only weights, adaptive and hardware-aware variants extend its flexibility to heterogeneous FL environments and edge devices, balancing accuracy with communication and computation efficiency.

\paragraph{Weight Clipping and Calibration}
Weight clipping and calibration are auxiliary mechanisms that operate within the training-stage and quantized-object dimensions, primarily supporting uniform scalar PTQ by stabilizing dynamic range and reducing quantization error by constraining and rescaling the numerical range of model parameters and activations.  These methods are particularly crucial when quantizing models for deployment on resource-constrained edge devices or in FL environments, where device heterogeneity and non-IID data distributions can cause severe quantization distortion. \textbf{Weight clipping} limits the dynamic range of weights by restricting them within a pre-defined threshold. Outlier values that fall outside this range are clipped, effectively reducing the quantization range and enabling finer granularity for frequently occurring small weights. The clipping process can be expressed as:
\begin{equation*}
w_c = \mathrm{clip}(w, -\alpha, \alpha) = 
\begin{cases}
-\alpha, & \text{if } w < -\alpha, \\
w, & \text{if } -\alpha \leq w \leq \alpha, \\
\alpha, & \text{if } w > \alpha,
\end{cases}
\end{equation*}
where $\alpha$ is the clipping threshold chosen to balance between preserving information and minimizing quantization noise.  Excessively large $\alpha$ values waste quantization levels on rare outliers, while overly small $\alpha$ values may remove significant signal information. To find an optimal $\alpha$, calibration techniques such as \textit{Kullback–Leibler Divergence} (KLD) minimization~\cite{migacz2017int8, choi2018pact} or \textit{Mean Squared Error} (MSE) optimization~\cite{nagel2020up} are applied. These methods estimate the threshold $\alpha$ that minimizes the divergence between the full-precision and quantized distributions. Formally, the optimal threshold is computed as $\alpha^* = \arg \min_{\alpha} D_{\mathrm{KL}}(P_{\text{float}}(x) \parallel P_{\text{quant}}(x; \alpha))$, where $P_{\text{float}}$ and $P_{\text{quant}}$ denote the distributions of the full-precision and quantized values, respectively.  

\textbf{Calibration} further refines quantization by determining scale factors and zero-points for each tensor based on representative calibration data. In practice, a small subset of data (typically 5–10\% of the training set) is used to estimate activation ranges. Calibration ensures that the quantization parameters reflect real operational data statistics, thereby minimizing activation saturation and quantization error~\cite{krishnamoorthi2018quantizing, Jacob2018CVPR}. In the context of \textit{FL}, weight clipping and calibration play a critical role in stabilizing communication and aggregation. Since each client may have a unique data distribution, applying global quantization thresholds can lead to biased model updates. Weight clipping with adaptive thresholds mitigates this by constraining local updates within consistent dynamic ranges, improving convergence stability. Calibration ensures that each client’s quantized updates align statistically with the global model, reducing aggregation bias. Techniques such as clipped uniform quantization in FL~\cite{bozorgasl2025communication} employ layer-wise clipping and calibration to balance communication efficiency with accuracy preservation. Overall, these methods significantly enhance quantized FL performance under heterogeneous and non-IID data conditions.

\begin{table*}[htbp]
\centering
\caption{Comparison of Weight Clipping and Calibration Methods in Neural Network Quantization and FL}
\resizebox{\textwidth}{!}{
\begin{tabular}{p{3.3cm} p{5.1cm} p{4.2cm} p{4.2cm} p{3.2cm}}
\hline
\textbf{Method/Contribution Area} & \textbf{Description/Approach} & \textbf{Advantages} & \textbf{Limitations} & \textbf{References} \\ 
\hline

\textbf{Fixed Threshold Clipping} &
Applies a manually selected static clipping threshold $\alpha$ to constrain weights or activations within $[-\alpha, \alpha]$. &
Simple and easy to implement; reduces the influence of extreme outliers. &
Suboptimal threshold may lead to under- or over-clipping; requires empirical tuning per layer.
 & \cite{krishnamoorthi2018quantizing} \\

\hline

\textbf{KLD-Based Calibration (Intel’s INT8 Calibration)} &
Determines optimal clipping thresholds by minimizing KLD between original and quantized distributions. &
Statistically optimal; minimizes quantization distortion; widely used in hardware frameworks like TensorRT and OpenVINO.
 & Computationally intensive; eequires representative calibration data.
& \cite{migacz2017int8, nvidia2021tensorrt} \\

\hline

\textbf{MSE-Based Clipping (Adaptive Rounding)} &
Selects $\alpha$ to minimize MSE between quantized and original weights; often combined with bias correction. &
Provides smooth error minimization; suitable for low-bit quantization; improves post-training accuracy without retraining.
 & Slightly increases calibration overhead; sensitive to layer-specific distributions.  & \cite{nagel2020up} \\

\hline

\textbf{PACT (Parameterised Clipping Activation)} &
Introduces a learnable clipping parameter $\alpha$ as part of training (QAT). & Learns optimal thresholds automatically; enhances model adaptability to low-bit precision.  &  Requires fine-tuning or retraining; increased training cost.  & \cite{choi2018pact} \\

\hline

\textbf{Layer-wise Clipped Quantization for FL} &
Employs layer-specific clipping and calibration across federated clients to ensure consistent quantization and reduce update variance. &
Enhances aggregation stability under non-IID data; reduces communication overhead with minimal accuracy loss.
 & Requires per-client calibration data; adds synchronization complexity in FL aggregation.
 & \cite{bozorgasl2025communication, Dubey2024uniform} \\

\hline
\end{tabular}
}
\label{tab:weight_clipping_calibration_comparison}
\vspace{-0.5cm}
\end{table*}

Table \ref{tab:weight_clipping_calibration_comparison} summarizes the key approaches and developments in weight clipping and calibration techniques for neural network quantization and FL. They function as range-control and parameter-estimation mechanisms that improve the robustness of scalar PTQ and QAT schemes. These methods aim to optimize the quantization range by constraining outliers and calibrating scaling factors to minimize quantization error. While fixed threshold and KLD-based calibration techniques provide simple yet effective post-training solutions, adaptive approaches such as MSE-based clipping and PACT introduce learnable or data-driven thresholds for improved precision. In FL contexts, layer-wise clipping and client-specific calibration enhance aggregation stability and reduce communication overhead under non-IID data distributions. Collectively, these methods highlight the evolution from static clipping toward adaptive and federated-aware quantization strategies that balance efficiency, robustness, and accuracy.

\paragraph{Quantization-Aware Training (QAT)}

QAT incorporates quantization effects directly into the training process to make neural networks robust to quantization noise during deployment. Unlike PTQ, which applies quantization after model training, QAT simulates low-precision arithmetic throughout forward and backward passes so that the model parameters are optimized with quantization constraints in mind~\cite{Jacob2018CVPR}. This integration allows QAT models to achieve accuracy levels close to their full-precision counterparts, even at ultra-low bit-widths such as 4-bit or 2-bit. Given a weight tensor $W$ and quantization function $Q(\cdot)$, the quantized weight used during training is expressed as $\tilde{W} = Q(W) = \mathrm{round}\!\left(\frac{W}{s}\right) \times s$, where $s$ is the scale factor determined per tensor, per-channel, or per-layer. The loss function used for training becomes $\mathcal{L}_{QAT} = \mathcal{L}_{\text{task}}(f(x; Q(W), Q(a))) + \lambda R(Q(W))$, where $\mathcal{L}_{\text{task}}$ is the primary task loss, $Q(a)$ denotes quantized activations, and $R(Q(W))$ is a regularization term (e.g., for stability or sparsity). Since quantization involves a non-differentiable rounding operation, gradients are approximated using the \textit{Straight-Through Estimator (STE)}~\cite{bengio2013ste}: $\frac{\partial \mathcal{L}}{\partial W} \approx \frac{\partial \mathcal{L}}{\partial Q(W)}$. This enables backpropagation through discrete quantization functions and ensures effective optimization.

\paragraph{Advanced Variants of QAT:} While standard QAT effectively models quantization noise during training, several advanced variants have emerged to enhance adaptability, convergence, and scalability in practical deployments. These methods focus on dynamically learning quantization parameters, improving gradient estimation, or enabling heterogeneous quantization settings in distributed environments such as FL.
\textit{Differentiable Quantization:} Differentiable quantization integrates quantization into the optimization process by relaxing discrete quantization operations into differentiable approximations. Instead of relying solely on the STE, this approach employs continuous surrogates such as the \textit{Gumbel-Softmax} or \textit{sigmoid-based approximations}~\cite{uhlich2019mixed, jang2017categorical}. This allows backpropagation through quantization levels, enabling end-to-end optimization of both weights and quantization thresholds. Differentiable quantization methods achieve smoother convergence and higher accuracy for low-bit quantization, particularly in transformer and vision models. \textit{Gradient-Based QAT (Learned Quantization Parameters):} In gradient-based QAT, both quantization scales and zero-points are treated as trainable parameters optimized using backpropagation~\cite{esser2020lsqplus, zhou2018doq}. This eliminates manual calibration and allows layer-wise or channel-wise bit-width adaptation. The gradient updates refine quantization parameters jointly with model weights, significantly improving model performance at extreme bit-widths (e.g., 2–4 bits). This approach forms the foundation for methods such as LSQ and LSQ+ that achieve near-floating-point accuracy without post-training fine-tuning. \textit{Adaptive Precision QAT:} Adaptive precision QAT dynamically adjusts bit-widths during training or across communication rounds in FL to optimize the trade-off between model accuracy and resource constraints. This technique employs layer sensitivity metrics or reinforcement learning agents to modify quantization levels as training progresses~\cite{yao2021adabits, Wang2019HAQ}. In federated settings, adaptive QAT enables heterogeneous clients to maintain different quantization precisions based on hardware capabilities or communication bandwidth~\cite{Zhang2021}. This flexibility enhances scalability while maintaining convergence consistency across clients. \textit{Federated Hybrid QAT:} Federated Hybrid QAT combines QAT with other compression techniques, such as pruning, mixed precision, or weight clustering—to further improve efficiency and robustness under non-IID data distributions. Clients apply QAT locally while the server aggregates models using adaptive precision control~\cite{Dubey2024uniform}. This hybridization helps correct quantization-induced drift and mitigates aggregation bias, allowing models to maintain accuracy even under asynchronous or partial participation scenarios.  

Overall, these advanced QAT variants extend traditional quantization-aware training toward differentiable, adaptive, and federated paradigms, enabling deep neural networks to operate efficiently in distributed and resource-constrained environments.

In the context of FL, QAT provides robustness against quantization-induced bias in distributed training. Since clients perform local updates with quantized weights and activations, QAT helps the global model adapt to device-level quantization errors, improving both convergence and accuracy~\cite{Danaee2022Biasquantization, bozorgasl2025communication}. Federated QAT frameworks, such as QA-FedAvg and QFL, integrate quantization simulation within each client’s local training loop, enabling effective training on heterogeneous devices. Furthermore, combining QAT with mixed-precision or clipping techniques enhances communication efficiency while maintaining near-floating-point performance~\cite{Dubey2024uniform, Zhang2021}. By quantizing updates before transmission, bandwidth consumption is significantly reduced. QAT-aware clients compensate for quantization errors during training, maintaining near full-precision accuracy. QAT allows heterogeneous clients (e.g., edge, IoT, or GPU devices) to train using precision levels suited to their computational capabilities. Incorporating bias correction and clipping during QAT reduces drift among client updates, ensuring more stable aggregation. QAT thus serves as a bridge between efficient model compression and accuracy preservation, especially in distributed edge and federated environments. Recent framework, \textit{QA-FedAvg}~\cite{Danaee2022Biasquantization} extends standard FL algorithms by incorporating QAT into local training. These frameworks differ primarily in how they quantize weights, gradients, or activations, such as \textbf{QA-FedAvg:} Combines QAT with FedAvg using bias correction to account for coarse quantization.  This framework demonstrate that QAT not only reduces model size and communication cost but also enhances robustness against client heterogeneity and gradient distortion in non-IID data environments.

\begin{algorithm}
\begin{algorithmic}[1]
\State \textbf{Input:} Global model $W^0$, quantizer $Q(\cdot)$, learning rate $\eta$, number of clients $K$.
\For{each communication round $t=1,2,\ldots,T$}
    \State Server sends current global model $W^t$ to selected clients.
    \For{each client $k=1$ to $K$}
        \State Initialize local model $W_k^t \leftarrow W^t$.
        \For{each local epoch}
            \State Quantize weights and activations: $\tilde{W}_k = Q(W_k^t)$.
            \State Compute gradients using STE: $\nabla \mathcal{L}(Q(W_k^t))$.
            \State Update weights: $W_k^t \leftarrow W_k^t - \eta \nabla \mathcal{L}$.
        \EndFor
        \State Send quantized model update $\Delta \tilde{W}_k^t$ to server.
    \EndFor
    \State Server aggregates updates: $W^{t+1} \leftarrow \frac{1}{K} \sum_k \Delta \tilde{W}_k^t$.
\EndFor
\State \textbf{Output:} Final global model $W^T$.

\end{algorithmic}
\caption{Quantization-Aware FL (QA-FL)}
\label{alg:qafl}
\end{algorithm}
In Quantization-Aware FL (QA-FL), clients locally train quantized models using fake quantization operations during both forward and backward passes, as shown in Algorithm~\ref{alg:qafl}. Quantization effects are simulated throughout training, and gradients are computed using the Straight-Through Estimator (STE), allowing clients to adapt their model weights to quantized representations. After each local training phase, the quantized updates are transmitted to the server, where aggregation (e.g., weighted averaging) is performed in full or mixed precision. This joint optimization ensures that both local and global models become robust to quantization-induced noise and communication compression. Overall, QAT-based FL frameworks combine quantization robustness with communication efficiency, offering a pathway to deploy deep learning at scale on resource-limited, heterogeneous edge networks.

\begin{table*}[htbp]
\centering
\caption{Unified comparison of QAT methods and QAT-based FL frameworks.}
\label{tab:qat_unified}
\resizebox{\textwidth}{!}{
\begin{tabular}{p{2.8cm} p{3.2cm} p{4.2cm} p{3.6cm} p{3.6cm} p{2.4cm}}
\hline
\textbf{Category} & \textbf{Method/Framework} & \textbf{Description} & \textbf{Advantages} & \textbf{Limitations} & \textbf{Key refs.} \\ 
\hline

Method & Standard QAT (Fake quant.) & Insert fake quantizers in forward pass and use STE in backward pass to train models robust to low-precision. & Simple; widely supported; preserves accuracy at modest bit-widths. & Extra local training cost; requires retraining/fine-tuning. & \cite{Jacob2018CVPR} \\

Method & PACT (Parameterized Clipping Activation) & Learnable clipping parameters for activations optimized jointly with weights. & Reduces activation saturation; automates bound selection. & Increases training complexity and hyperparameter tuning. & \cite{choi2018pact} \\

Method & LSQ / LSQ+ (Learned Step Size Quant.) & Treat scale / zero-point / step sizes as trainable parameters (end-to-end). & Near full-precision accuracy at low bits; joint weight/scale optimization. & Sensitive to initialization; slightly slower convergence. & \cite{esser2020lsqplus} \\

Method & QDrop (Quantization Dropout) & Randomly skip quantization for subsets of weights/activations during training. & Improves generalization under extreme quantization; reduces overfitting. & Adds stochastic variance; tuning required. & \cite{qdrop2022} \\

Method & Differentiable Quantization & Replace hard rounding with continuous surrogates (e.g., Gumbel-Softmax) for better gradients. & Smoother optimization; can outperform STE in some settings. & Higher computational overhead; approximation mismatch. & \cite{uhlich2019mixed, jang2017categorical} \\

Method & Adaptive Precision QAT & Dynamically adjust per-layer / per-client bit-widths (sensitivity metrics or RL). & Hardware-aware; balances accuracy vs. latency/energy; ideal for heterogeneous clients. & Scheduling complexity; metadata and synchronization overhead. & \cite{Wang2019HAQ, yao2021adabits, Zhang2021} \\

Method & Federated Hybrid QAT & Combine QAT with pruning, clustering, mixed-precision or other compression. & Strong compression + robustness to non-IID data; flexible trade-offs. & Aggregation complexity; requires client coordination/sync. & \cite{Dubey2024uniform} \\

\hline

Framework & QA-FedAvg & FedAvg extended with local QAT and bias compensation to reduce quantization bias. & Improves convergence under coarse quantization; retains accuracy. & Requires local retraining; extra client compute. & \cite{Danaee2022Biasquantization} \\



Framework & FedDQ (Differentiable QAT for FL) & Applies differentiable quantizers to clients for smoother gradient propagation. & Reduces quantization drift; improves stability and convergence. & Computationally demanding; sensitive to surrogate choice. & \cite{uhlich2019mixed, jang2017categorical} \\

Framework & FedMixQAT / Adaptive-Precision FL & Enables dynamic client bit-widths based on device capacity and network. & Reduces communication adaptively; improves device-specific efficiency. & Bit-width synchronization and fairness issues; scheduling overhead. & \cite{Dubey2024uniform} \\

\hline
\end{tabular}
}
\vspace{-0.5cm}
\end{table*}

 Table \ref{tab:qat_unified} provides a unified comparison of major QAT methods and QAT-based FL frameworks. It consolidates standard approaches such as fake quantization, PACT, and LSQ with advanced variants including differentiable, adaptive, and hybrid precision quantization. The table also summarizes key FL frameworks, such as QA-FedAvg, FedDQ, and FedMixQAT-highlighting how QAT has evolved from basic quantization simulation to adaptive, device-aware, and communication-efficient paradigms. Collectively, these methods demonstrate that QAT not only preserves model accuracy under low-bit regimes but also enhances robustness and scalability in heterogeneous FL environments.

\paragraph{Mixed Precision Quantization}
Mixed Precision Quantization (MPQ) is an advanced quantization strategy that assigns different bit-widths to different layers, channels, or tensors within a neural network, instead of using a uniform bit-width across the entire model. It can be implemented under either PTQ or QAT without changing the underlying encoding structure. The key idea is to apply higher precision (e.g., 16-bit or 8-bit) to layers or parameters that are more sensitive to quantization noise, while using lower precision (e.g., 4-bit or 2-bit) for less critical layers. This adaptive allocation of bit-widths achieves a better trade-off between model accuracy and computational efficiency compared to fixed uniform quantization~\cite{Dong2019HAWQ, Wang2019HAQ, Banner2019uniform, choi2021mpqsurvey}. Formally, let each layer $l$ in the model be quantized using a distinct bit-width $b_l$, with the objective of minimizing a global cost function that balances accuracy and resource consumption as $\min_{\{b_l\}} \mathcal{L}_{\text{task}}(Q(W; \{b_l\})) + \lambda \sum_{l} \mathrm{Cost}(b_l)$, where $\mathcal{L}_{\text{task}}$ denotes the task-specific loss (e.g., classification loss), $Q(W; \{b_l\})$ represents the quantized model under bit-width configuration $\{b_l\}$, and $\mathrm{Cost}(b_l)$ models hardware or communication cost constraints. The regularization term $\lambda$ controls the trade-off between efficiency and accuracy. MPQ can be implemented using analytical, reinforcement learning (RL), or differentiable search methods. Analytical approaches rely on sensitivity analysis or Hessian-based metrics~\cite{Dong2019HAWQ} to estimate how quantization affects each layer’s loss, while RL-based methods~\cite{Wang2019HAQ} automatically learn optimal bit-width assignments through feedback from performance and latency metrics. More recent differentiable methods, such as AdaQuant and HAQ-V2~\cite{yao2021adabits}, integrate quantization into the training process, enabling gradient-based optimization over bit-widths.

In the context of \textit{FL}, MPQ provides a flexible mechanism for adapting models to heterogeneous client hardware. Clients with powerful GPUs can operate with higher bit-widths, whereas edge devices or IoT clients can use lower precision to reduce computational load and communication cost. This heterogeneous deployment improves the inclusivity and scalability of FL systems. Moreover, adaptive MPQ can dynamically adjust precision across training rounds, improving energy efficiency and communication efficiency without sacrificing global accuracy~\cite{Dubey2024uniform}. While MPQ offers considerable advantages in balancing efficiency and precision, it introduces complexity in optimization and calibration. The need to manage multiple quantization scales and per-layer configurations can increase the implementation burden, particularly in federated environments where synchronization between clients is required. Nonetheless, MPQ remains one of the most promising directions for achieving hardware-aware and communication-efficient FL.

\begin{table*}[htbp]
\centering
\caption{Comparison of MPQ Methods in Neural Network Compression and FL}
\resizebox{\textwidth}{!}{
\begin{tabular}{p{3.3cm} p{5.1cm} p{4.2cm} p{4.2cm} p{3.2cm}}
\hline
\textbf{Method/Contribution Area} & \textbf{Description/Approach} & \textbf{Advantages} & \textbf{Limitations} & \textbf{Key References} \\ 
\hline

\textbf{HAQ (Hardware-Aware Quantization)} &
Uses reinforcement learning to determine per-layer bit-widths based on hardware latency and energy constraints. &

   Hardware-aware; automatically balances accuracy and latency; Suitable for diverse deployment environments.
 &

   Requires reinforcement learning search, increasing computational overhead; Dependent on hardware simulation accuracy.
 &
\cite{Wang2019HAQ} \\

\hline

\textbf{HAWQ (Hessian-Aware Quantization)} &
Uses second-order sensitivity analysis (Hessian information) to assign per-layer bit-widths. Layers with higher sensitivity are allocated higher precision. &

   Provides theoretically grounded precision allocation;
   Outperforms uniform quantization with minimal accuracy drop.
 &

    Requires computation of Hessian approximation;
 Limited scalability for very large models.
 &
\cite{Dong2019HAWQ} \\

\hline

\textbf{AdaBits / AdaQuant} &
Adopts differentiable optimization to learn bit-widths and quantized weights during training jointly. &    Fully differentiable and efficient;
   Produces optimal bit allocation dynamically during training.  &   Requires quantization-aware training;
   Computationally expensive for large networks.  & \cite{yao2021adabits} \\

\hline

\textbf{Mixed Precision Quantization for FL (MPQ-FL)} &
Extends MPQ to FL by allowing clients to adopt variable bit-widths based on device capacity and network conditions. &

 Adapts precision to device heterogeneity; Reduces communication bandwidth and energy consumption.
 &

    Requires synchronization and calibration between clients; Possible divergence due to heterogeneous quantization settings.
 &
\cite{Dubey2024uniform} \\

\hline

\textbf{Layer-Wise Adaptive MPQ} &
Employs per-layer or per-channel adaptive precision using sensitivity analysis and model calibration. &

   Fine-grained optimization of bit-width allocation; Maintains high accuracy with significant compression.
 &

 Increased calibration complexity; Not yet standardized across hardware platforms.
 &
\cite{Banner2019uniform, choi2021mpqsurvey} \\

\hline
\end{tabular}
}
\label{tab:mixed_precision_quantization_comparison}
\vspace{-0.5cm}
\end{table*}

Table \ref{tab:mixed_precision_quantization_comparison} summarizes major MPQ methods developed for neural network compression and FL. These approaches dynamically allocate precision across layers or clients to balance accuracy, energy efficiency, and communication cost. While methods such as HAWQ and HAQ employ Hessian-based or reinforcement learning techniques to determine bit-widths, federated extensions like MPQ-FL adapt quantization settings to client hardware heterogeneity. The evolution of MPQ highlights a shift from static, uniform quantization toward adaptive, device-aware, and communication-efficient strategies for large-scale distributed learning.

Quantization in FL can be applied to various entities in the training process, primarily including model parameters, gradients, and activations. 
\paragraph{Parameter Quantization:} In parameter quantization, the global model parameters are encoded into lower-precision formats before being transmitted to clients. Each weight $w_i$ is mapped to a discrete value $Q(w_i)$ according to $Q(w_i) = \mathrm{round}\!\left(\frac{w_i}{s}\right) \times s$, where $s$ is the quantization step size, defined as $s = \frac{w_{\max} - w_{\min}}{2^b - 1}$ for a $b$-bit representation~\cite{Jacob2018CVPR}. This reduces the model’s memory footprint and communication cost linearly with respect to $b$ while maintaining the approximate structure of the global model, making it ideal for low-bandwidth or energy-constrained edge devices~\cite{Reisizadeh2020, Shen2020, Jhunjhunwala2021}. In particular, Reisizadeh et al.~\cite{Reisizadeh2020} introduced FedPAQ, which combines periodic averaging with parameter quantization to lower communication cost while maintaining convergence guarantees. Similarly, Shen et al.~\cite{Shen2020} proposed QFNN, a bit-limited model quantization technique applied directly to network parameters before uplink communication. Jhunjhunwala et al. ~\cite{Jhunjhunwala2021} extended this idea with AdaQuantFL, adaptively changing quantization bit-widths during training to balance convergence and bandwidth efficiency. More recently, Ni et al. ~\cite{Ni2024_a} integrated parameter quantization with energy-aware scheduling, showing significant gains in sustainability for edge devices. Overall, parameter quantization remains the backbone of FL compression methods, serving as the foundation for more advanced schemes such as gradient quantization, mixed-precision quantization, and vector quantization.

\paragraph{Gradient Quantization:} Gradient quantization aims to compress updates $\nabla W_k^t$ before communication to the server. The quantized gradient $\tilde{\nabla W_k^t}$ is given as $\tilde{\nabla W_k^t} = Q(\nabla W_k^t) = s \cdot \mathrm{round}\!\left(\frac{\nabla W_k^t}{s}\right)$, where $Q(\cdot)$ is a stochastic or deterministic quantization function~\cite{Alistarh2017uniform}. Stochastic quantization preserves unbiasedness, i.e., $\mathbb{E}[Q(x)] = x$, ensuring that the global gradient estimator remains consistent as $\mathbb{E}\left[\tilde{\nabla W_k^t}\right] = \nabla W_k^t$. Gradient quantization has emerged as one of the most effective strategies to alleviate the communication bottleneck in FL by compressing model updates before transmission. Unlike parameter quantization, which targets static model weights, gradient quantization focuses on dynamically exchanged gradients, enabling significant bandwidth reduction during each training round. Early studies such as QSGD~\cite{Alistarh2017uniform} established the theoretical foundations for unbiased stochastic quantization, while subsequent works including FedPAQ~\cite{Reisizadeh2020} and LAQ~\cite{Sun2020} demonstrated the benefits of combining quantization with periodic averaging and lazy aggregation. More recent frameworks like DAdaQuant~\cite{Honig2022}, AdaGQ~\cite{Liu2023_a}, and FedVQCS~\cite{Oh2024} introduced adaptive, client-specific, and vector-based quantization to further balance accuracy, convergence, and communication cost. 

\paragraph{Activation and Intermediate Quantization:} Activation and intermediate quantization aim to reduce the memory footprint and computational cost of local training in FL by compressing the activation tensors and intermediate feature maps generated during forward propagation. Unlike parameter or gradient quantization, which focus on communication efficiency, activation quantization primarily targets on-device efficiency—making FL feasible on mobile and IoT devices with limited memory and power budgets~\cite{krishnamoorthi2018quantizing, Shen2020, choi2018pact}.  Formally, consider an activation tensor $\mathbf{a}_l \in \mathbb{R}^{n_l}$ produced at layer $l$ during forward propagation. 
The quantized activation $\hat{\mathbf{a}}_l$ is obtained by mapping $\mathbf{a}_l$ to a discrete set of representable values based on a quantization step size $s_l$, computed from the activation range:
$\hat{\mathbf{a}}_l = Q(\mathbf{a}_l) = s_l \cdot \mathrm{round}\!\left(\frac{\mathbf{a}_l}{s_l}\right),
\quad 
s_l = \frac{\max(\mathbf{a}_l) - \min(\mathbf{a}_l)}{2^{b_l} - 1}$, where $b_l$ denotes the bit-width used for quantizing activations at layer $l$.  

In practice, the quantization process can also be expressed in integer form by defining the integer activation representation $\mathbf{a}_l^q$ and its dequantized approximation $\hat{\mathbf{a}}_l$ using $\mathbf{a}_l^q = \mathrm{clip}\!\left(\mathrm{round}\!\left(\frac{\mathbf{a}_l}{s_l}\right) + z_l, \, 0, \, 2^{b_l}-1\right),
\quad
\hat{\mathbf{a}}_l = s_l \cdot (\mathbf{a}_l^q - z_l)$, where $z_l$ is the zero-point that ensures non-negative quantized representations (used in asymmetric quantization).  

This process effectively replaces floating-point activations with low-bit integer approximations (e.g., INT8 or INT4), which reduces both the memory access cost and the energy required for matrix multiplications on embedded hardware. However, as deeper layers tend to amplify quantization noise, proper calibration and clipping techniques—such as PACT~\cite{choi2018pact} or per-layer scaling~\cite{Jacob2018CVPR}—are essential to maintain convergence and accuracy.

In the context of FL, activation and intermediate quantization are particularly effective for on-device training and privacy-preserving inference. For instance, Shen \textit{et al.}~\cite{Shen2020} introduced the \textit{Quantized Federated Neural Network} (QFNN), which quantizes both parameters and activations to reduce client-side memory usage while preserving model accuracy. Similarly, Feng \textit{et al.}~\cite{Feng2022} developed a fully quantized activation (FAQ) scheme for federated edge devices, quantizing both weights and activations to INT8 for end-to-end efficiency. 

\textit{Intermediate quantization}, a related technique, compresses intermediate feature maps exchanged between layers or between clients and servers in split or hierarchical FL setups~\cite{Thapa2020SplitFedWF, Bakhtiarnia2022dynamicsplitcomputingefficient}. Let $\mathbf{f}_l$ represent the feature map at an intermediate layer $l$ transmitted from the client to the server. The quantized feature map $\hat{\mathbf{f}}_l$ is computed as $\hat{\mathbf{f}}_l = s_f \cdot \mathrm{round}\!\left(\frac{\mathbf{f}_l}{s_f}\right),
\quad
s_f = \frac{\max(\mathbf{f}_l) - \min(\mathbf{f}_l)}{2^{b_f}-1}$, where $b_f$ is the bit-width for feature quantization. By transmitting $\hat{\mathbf{f}}_l$ instead of $\mathbf{f}_l$, clients achieve substantial communication savings during forward-backward synchronization.  

Although activation and intermediate quantization significantly enhance computational and communication efficiency, they introduce challenges such as activation saturation, quantization noise propagation, and accuracy degradation in deep or non-IID federated settings. To mitigate these issues, adaptive calibration, per-layer clipping, and mixed-precision quantization strategies have been proposed~\cite{choi2018pact, Jacob2018CVPR}. Overall, activation and intermediate quantization form a critical bridge between hardware-efficient FL deployment and energy-aware distributed intelligence.

\paragraph{Model Update Quantization:} Alternatively, clients transmit quantized model updates $\Delta W_k^t = W_k^t - W^{t-1}$ rather than full model parameters, i.e., $Q(\Delta W_k^t) = s \cdot \mathrm{round}\!\left(\frac{\Delta W_k^t}{s}\right)$, which further improves communication efficiency in synchronous aggregation frameworks~\cite{Reisizadeh2020}. Model update quantization has emerged as a practical strategy to improve communication efficiency in FL. Instead of transmitting complete model parameters, clients communicate quantized updates $\Delta W_k^t = W_k^t - W^{t-1}$, which substantially reduces bandwidth requirements during each aggregation round. Reisizadeh \textit{et al.}~\cite{Reisizadeh2020} introduced FedPAQ, combining periodic averaging with uniform quantization, while Sun \textit{et al.}~\cite{Sun2020} proposed LAQ, leveraging lazy aggregation and innovation-based quantization. Further, Amiri and Gündüz~\cite{amiri2020federatedlearningquantizedglobal} explored quantized model transmission over wireless channels, and Elkordy and Avestimehr~\cite{Elkordy2022} enhanced this approach through secure and heterogeneous quantization under variable bandwidths. Recent adaptive approaches such as AdaGQ~\cite{Liu2023_a} dynamically adjust quantization precision per client based on gradient norms and device capabilities. 


\paragraph{Uniform Quantization}
Uniform quantization is one of the most widely implemented quantization strategies in FL due to its simplicity, hardware efficiency, and ease of integration into existing communication and computation pipelines. The fundamental idea is to represent model parameters or gradients with equally spaced discrete values by dividing the numerical range $[x_{\min}, x_{\max}]$ into fixed-size intervals. Each real value $x$ is approximated by its nearest quantized level according to $x_q = \mathrm{round}\!\left(\frac{x - x_{\min}}{s}\right) \times s + x_{\min}, \quad s = \frac{x_{\max} - x_{\min}}{2^{b} - 1}$, where $b$ denotes the bit-width, and $s$ is the quantization step size. In the context of FL, uniform quantization significantly reduces communication overhead by compressing model updates exchanged between clients and the server, while maintaining competitive accuracy. However, a key limitation arises from \textit{outlier values}-large deviations in model weights or gradients, which stretch the quantization range and thus lower precision for frequently occurring small values. This is particularly challenging in non-IID client settings, where data heterogeneity causes wide dynamic variations in parameter distributions. Recent research has refined the uniform quantization paradigm through clipping and stochastic extensions. For instance, Bozorgasl and Chen~\cite{bozorgasl2025communication} proposed clipped uniform quantization that adaptively determines optimal clipping thresholds to minimize quantization noise while maintaining accuracy. Their method introduces stochastic quantization, which adds controlled randomness to reduce bias and enhance robustness during aggregation. Similarly, Dubey and Kumar~\cite{Dubey2024uniform} compared uniform quantization with post-training and quantization-aware methods, showing that uniform quantization achieves substantial memory reduction (up to 75\%) and inference acceleration (2-4$\times$) with minimal accuracy loss (1-2\%). Overall, uniform quantization remains a baseline technique for communication-efficient FL, often serving as a foundation for more advanced quantizers (e.g., per-layer, non-uniform, or clipped methods). Its combination of simplicity, interpretability, and hardware compatibility (e.g., INT8) makes it indispensable for scalable, privacy-preserving federated systems deployed on heterogeneous edge devices.

\begin{table*}[htbp]
\centering
\caption{Comparison of Uniform Quantization Variants in FL}
\vspace{-0.5cm}
\resizebox{\textwidth}{!}{
\begin{tabular}{p{3.2cm} p{5.3cm} p{4.2cm} p{4.2cm} p{3.2cm}}
\hline
\textbf{Contribution Area} & \textbf{Description/Method} & \textbf{Advantages} & \textbf{Limitations} & \textbf{References} \\ 
\hline
\textbf{Standard Uniform Quantization} & 
Fixed step-size quantization with evenly spaced bins across $[x_{\min}, x_{\max}]$. &
Simple, hardware-efficient (supported by TensorFlow Lite, TensorRT), low computational cost. &
Sensitive to outliers; reduced precision for non-uniform data distributions. &
\cite{Dubey2024uniform, Jacob2018CVPR} \\

\textbf{Clipped Uniform Quantization} &
Limits outlier influence by clipping values outside a threshold before quantization. &
Reduces quantization noise; adaptive control of range; improves robustness. &
May discard valuable information if clipping is too aggressive. &
\cite{bozorgasl2025communication, Banner2019uniform} \\

\textbf{Stochastic Uniform Quantization} &
Adds uniform random noise before quantization to reduce bias (also called dithered quantization). &
Reduces deterministic quantization bias; improves convergence and generalization. &
Adds randomness and slight variance in updates. &
\cite{bozorgasl2025communication, Alistarh2017uniform} \\

\textbf{Adaptive/Per-Layer Uniform Quantization} &
Applies distinct quantization scales or bit-widths for different layers or clients. &
Balances accuracy and efficiency; mitigates data heterogeneity in FL. &
Increased computational complexity and calibration cost. &
\cite{Dubey2024uniform, Park2018uniform} \\

\hline
\end{tabular}
}
\label{tab:uniform_quantization_comparison}
\vspace{-0.5cm}
\end{table*}
Table \ref{tab:uniform_quantization_comparison} provides a comparative summary of the major variants of uniform quantization employed either in NN or FL. The table highlights their core methodologies, advantages, and limitations, emphasizing how each successive refinement, ranging from standard fixed-step quantization to clipped, stochastic, and adaptive/per-layer approaches, enhances communication efficiency and model robustness. Standard uniform quantization remains the simplest and most hardware-friendly technique but struggles with outlier sensitivity and non-uniform data. Clipped and stochastic variants mitigate quantization noise and bias, while adaptive or per-layer strategies introduce flexibility across model layers or clients, balancing accuracy with efficiency in heterogeneous FL environments. Collectively, these methods illustrate the evolution of uniform quantization from basic discretization toward adaptive, context-aware compression techniques that better suit the challenges of decentralized and resource-constrained learning systems.

\paragraph{Non-uniform Quantization:} Non-uniform quantization extends the concept of fixed-step quantization by allocating quantization levels unevenly across the value range to better match the statistical distribution of model parameters or activations. Unlike uniform quantization, which assigns equal spacing between quantization bins, non-uniform schemes use variable step sizes, often smaller near dense regions (e.g., around zero) and larger toward the tails of the distribution, thereby minimizing quantization error and improving low-bit precision \cite{Sun2016, Hao2023}. Methods such as Lloyd–Max quantization, logarithmic scaling, and learned non-uniform step sizes adaptively fit quantization intervals to the data distribution, achieving better trade-offs between accuracy and compression \cite{Gongyo2024ACCV, Yvince2023}. Techniques like Additive Powers-of-Two (APoT) \cite{Hao2023} and PowerQuant \cite{Yvince2023} offer hardware-friendly implementations with non-uniform discretization. In FL, distribution-aware non-uniform quantization further enhances communication efficiency by tailoring quantization to client-specific data distributions, as demonstrated in approaches like FedWSQ~\cite{Kim2025} and Non-QuanFL \cite{Chen2024nqfl}. However, these adaptive schemes introduce additional computational complexity and calibration overhead compared to simpler uniform quantization methods.
\begin{table*}[h]

\centering
\caption{Comparison of Non-Uniform Quantization Methods in FL and Neural Network Compression}
\vspace{-0.5cm}
\resizebox{\textwidth}{!}{
\begin{tabular}{p{4.0cm} p{5.1cm} p{4.2cm} p{4.2cm} p{1.5cm}}
\hline
\textbf{Method/Contribution Area} & \textbf{Description/Approach} & \textbf{Advantages} & \textbf{Limitations} & \textbf{References} \\ 
\hline

\textbf{Intra-Layer Non-Uniform Quantization} &
Applies equal-distance and K-means-based non-uniform quantization to weights and activations at the layer level. &
Captures layer-wise value distribution; reduces quantization error compared to uniform schemes. &
Increased computational complexity during calibration and retraining. &
\cite{Sun2016} \\

\textbf{Learning Non-Uniform Step Sizes} &
Learns different quantization step sizes for each layer or channel via optimization. &
Adaptively aligns quantization intervals with data distribution; minimizes distortion. &
Requires gradient-based optimization; adds training overhead. &
\cite{Gongyo2024ACCV} \\

\textbf{Additive Powers-of-Two Quantization (APoT)} &
Represents quantized weights as sums of powers of two, achieving non-uniform discretization with hardware-friendly operations. &
Combines non-uniform precision with efficient bit-shift hardware implementation. &
Limited flexibility for arbitrary value distributions; fixed representation granularity. &
\cite{Hao2023} \\

\textbf{PowerQuant} &
Performs data-free automorphism search to derive optimal non-uniform mapping functions (power-based scaling). &
Enables post-training quantization without retraining; robust across datasets. &
Requires heuristic search; moderate computational overhead for transformation learning. &
\cite{Yvince2023} \\

\textbf{CoNLoCNN (Correlation + Non-Uniform Quantization)} &
Introduces correlation-based error compensation for low-precision CNNs with non-uniform quantization. &
Improves low-bit performance while maintaining structural sparsity and model accuracy. &
Specific to CNNs; limited generalization to non-vision models. &
\cite{Hanif2022} \\

\textbf{FedWSQ (Distribution-Aware Non-Uniform Quantization)} &
Implements non-uniform quantization in FL using client-specific distributions with adaptive weight standardization. &
Enhances model accuracy under heterogeneous client data distributions. &
Requires client-side computation for distribution modeling; potential privacy risks. &
\cite{Kim2025} \\

\textbf{Non-QuanFL (Non-Uniform Quantized FL)} &
Applies Lloyd-Max-based non-uniform quantization to gradients and updates in FL to reduce communication load. &
Efficient communication; balances accuracy and compression adaptively. &
May suffer from added quantization bias if distribution is not well estimated. &
\cite{Chen2024nqfl} \\

\textbf{LM-DFL (Lloyd-Max Decentralized FL)} &
Uses Lloyd-Max non-uniform quantization in decentralized FL for dynamic distortion minimization. &
Improves robustness in peer-to-peer communication; adaptive precision per device. &
Increased algorithmic complexity for dynamic quantizer updates. &
\cite{chen2024communication} \\

\textbf{Optimization-Based Non-Uniform FL (GQFedWAvg)} &
Supports both uniform and non-uniform quantization options, optimized via gradient-based averaging in FL. &
Hybrid flexibility; adapts quantization to available device resources and bandwidth. &
High tuning complexity; may require hyperparameter adjustment per client. &
\cite{Li2024GQFedWAvg} \\

\hline
\end{tabular}
}
\label{tab:nonuniform_quantization_comparison}
\vspace{-0.5cm}
\end{table*}

Table \ref{tab:nonuniform_quantization_comparison} summarizes the primary non-uniform quantization techniques used in neural networks and FL. These approaches introduce adaptive quantization levels that better fit data or weight distributions, minimizing quantization noise and preserving model accuracy in heterogeneous environments. While methods like APoT and PowerQuant achieve hardware-friendly implementations, FL-specific schemes such as FedWSQ and Non-QuanFL extend non-uniform quantization to communication-efficient, privacy-preserving decentralized learning frameworks.

\paragraph{Vector Quantization}
Vector Quantization (VQ) groups model parameters or gradients into small fixed-size vectors and replaces each vector with the index of the closest codeword from a pre-defined \textit{codebook} (dictionary). Instead of transmitting every value independently, as in scalar quantization, clients send indices referring to codewords, thereby drastically reducing communication overhead. Mathematically, the quantization process can be expressed as $q(\mathbf{x}) = \mathbf{c}_k, \quad \text{where } k = \arg\min_{j} \|\mathbf{x} - \mathbf{c}_j\|_2$, where $\mathbf{x} \in \mathbb{R}^d$ is a vectorized group of model parameters or gradients,     $\mathcal{C} = \{\mathbf{c}_1, \mathbf{c}_2, \ldots, \mathbf{c}_K\}$ denotes the shared codebook containing $K$ representative codewords, and $\mathbf{c}_k$ is the codeword that minimizes the Euclidean distortion $\|\mathbf{x} - \mathbf{c}_j\|_2$. During model aggregation, the server reconstructs the approximate updates using $\hat{\mathbf{x}} = \mathbf{c}_k$, where the transmitted index $k$ corresponds to the nearest codeword in the codebook. VQ achieves $\mathcal{O}(\log_2 K)$ communication cost per vector group instead of $\mathcal{O}(d)$ for raw transmission, making it highly suitable for large-scale FL systems involving thousands of clients. By capturing correlations among grouped parameters, vector quantization enables a better rate–distortion trade-off compared to scalar quantization, providing a balance between accuracy and compression efficiency in communication-constrained federated settings. While standard VQ provides a compact representation of model updates through shared codebooks, its scalability and reconstruction accuracy can be further enhanced by advanced variants such as \textit{Product Quantization (PQ)} and \textit{Lattice Quantization (LQ)} approaches extend VQ by improving the trade-off between compression ratio, distortion, and computational efficiency, making them particularly suitable for large-scale and heterogeneous FL environments.

\paragraph{Product Quantization:} Product Quantization divides a high-dimensional vector (e.g., gradient or model update) into multiple lower-dimensional subvectors, each quantized independently using its own sub-codebook~\cite{Jegou2011, Yang2024productquantization}. Formally, given a parameter vector $\mathbf{x} \in \mathbb{R}^d$, it is decomposed into $M$ sub-vectors as $\mathbf{x} = [\mathbf{x}_1, \mathbf{x}_2, \ldots, \mathbf{x}_M]$, where each $\mathbf{x}_m \in \mathbb{R}^{d/M}$ is quantized using its corresponding sub-codebook $\mathcal{C}_m = \{\mathbf{c}_{m,1}, \ldots, \mathbf{c}_{m,K}\}$.  
The quantization function is defined as 
\begin{equation*}
\begin{split}
Q_{PQ}(\mathbf{x}) = [q_1(\mathbf{x}_1), q_2(\mathbf{x}_2), \ldots, q_M(\mathbf{x}_M)], 
\\
q_m(\mathbf{x}_m) = \mathbf{c}_{m, k^*}, \;
k^* = \arg\min_k \|\mathbf{x}_m - \mathbf{c}_{m,k}\|_2.
\end{split}
\end{equation*}
The reconstructed vector at the server is given by $\hat{\mathbf{x}} = [\mathbf{c}_{1,k_1}, \mathbf{c}_{2,k_2}, \ldots, \mathbf{c}_{M,k_M}]$.

This decomposition enables exponential compression efficiency, as the number of possible reconstructed vectors becomes $K^M$, while storage and communication scale only with $M \log_2 K$.  
PQ is highly effective in FL, where clients can independently quantize sub-vectors using local sub-codebooks, and the server performs codeword index aggregation. Yang \textit{et al.}~\cite{Yang2024productquantization} introduced \textbf{FedMPQ}, a secure and communication-efficient FL framework that employs multi-codebook PQ combined with residual pruning and secure aggregation. Each client transmits compressed updates, pseudo-centroids, and residuals, while the server constructs global shared codebooks in a trusted execution environment (TEE) to ensure privacy.  
PQ achieves up to $90\%$ communication reduction with minimal accuracy degradation, outperforming scalar and standard vector quantization methods.

\paragraph{Lattice Quantization:} Lattice Quantization encodes model updates onto discrete lattice points within Euclidean space, enabling highly structured and computationally efficient quantization. Given a lattice $\Lambda \subset \mathbb{R}^d$, the quantized vector $\hat{\mathbf{x}}$ is the nearest lattice point to $\mathbf{x}$ as 
$Q_{LQ}(\mathbf{x}) = \arg\min_{\lambda \in \Lambda} \|\mathbf{x} - \lambda\|_2$. A simple example is the cubic lattice $\Lambda = s \mathbb{Z}^d$, where each coordinate of $\mathbf{x}$ is rounded to the nearest multiple of the step size $s$. More advanced constructions such as the \textit{Voronoi lattice} or \textit{E8 lattice}~\cite{Conway1999sphere} minimize quantization error through geometric packing optimization.

In FL, Shlezinger \textit{et al.}~\cite{Shlezinger2020_a} and Chen \textit{et al.}~\cite{Chen2021communication} applied lattice-based quantization to encode model updates and gradient information efficiently. LQ allows unbiased gradient reconstruction while achieving near-optimal rate–distortion trade-offs. The \textbf{JoPEQ} framework~\cite{Lang2023} extended this by combining lattice quantization with privacy-preserving additive noise to satisfy local differential privacy (LDP) constraints. Similarly, \textbf{Universal Vector Quantization (UVQ)}~\cite{Chen2021communication} used random lattice structures to achieve communication-efficient and task-agnostic compression. Lattice quantization is particularly advantageous for \textit{wireless FL} and \textit{over-the-air aggregation}, where structured encoding enables low-complexity decoding at the parameter server~\cite{amiri2020federatedlearningquantizedglobal, Shlezinger2020_a}. However, challenges such as high encoding complexity for large-scale models and synchronization of random lattice parameters among clients remain open research directions. Product and lattice quantization thus represent advanced directions in model compression for FL, jointly optimizing communication efficiency, privacy, and accuracy. While PQ focuses on modular compression through sub-vector decomposition, LQ provides geometric regularity and efficient reconstruction under structured encoding. Future research trends aim to integrate these techniques with adaptive bit-allocation, privacy guarantees, and mixed-precision optimization to further enhance federated scalability.


Quantization schemes can be further classified based on the precision levels and the quantization step adaptation strategy.

\paragraph{Fixed-Point Quantization.}
Fixed-point quantization (FPQ) is a widely used and hardware-compatible approach for reducing the precision of model parameters and gradients in federated learning (FL). FPQ represents real-valued variables using a fixed bit-width \(b\) (typically 8 or 16 bits) and integer arithmetic, enabling efficient computation and reduced memory footprint on CPUs, GPUs, and edge accelerators. Given a scaling factor \(s\), a real-valued parameter \(x\) is quantized as $x_q = \mathrm{clip}\!\left(\mathrm{round}\!\left(\tfrac{x}{s}\right), -2^{b-1}, 2^{b-1}-1\right), \quad 
\hat{x} = s \cdot x_q ,$ where the \textit{clip} operation enforces the representable dynamic range. Owing to its simplicity and low overhead, FPQ has been widely adopted in practical inference and training pipelines~\cite{krishnamoorthi2018quantizing,Jacob2018CVPR}.

Despite these advantages, standard FPQ typically employs static scaling factors, which can be suboptimal in FL settings characterized by non-IID data and heterogeneous client updates. Fixed scaling may induce quantization bias, saturation, or degraded convergence when parameter or gradient ranges vary significantly across clients or communication rounds. To mitigate these effects, recent work has explored adaptive fixed-point quantization schemes that dynamically adjust scaling factors based on observed data statistics. In particular, dynamic fixed-point quantization (DFPQ) computes the scaling factor at communication round \(t\) as $s_t = \tfrac{\max(\mathbf{x}_t) - \min(\mathbf{x}_t)}{2^b - 1}$, allowing quantization precision to adapt to evolving parameter or gradient distributions. Such dynamic recalibration has been shown to stabilize convergence under heterogeneous updates and to be especially effective for low-bit (\(\leq 8\)-bit) quantization. For example, Reisizadeh \textit{et al.}~\cite{Reisizadeh2020} incorporated adaptive fixed-point quantization into FedPAQ, while Honig \textit{et al.}~\cite{Honig2022} proposed doubly adaptive schemes combining dynamic scaling and stochastic rounding to improve robustness under non-IID data.

\paragraph{Asymmetric Quantization:} In contrast to symmetric quantization, which assumes zero-centered distributions, asymmetric quantization introduces a \textit{zero-point offset} $z$ to shift the representable integer range, enabling more accurate mapping of non-zero-mean tensors~\cite{Jacob2018CVPR, krishnamoorthi2018quantizing}.  This technique is widely used in hardware implementations (e.g., TensorRT, TFLite) to minimize bias and improve precision under uneven value distributions. Given a tensor $\mathbf{x}$ and bit-width $b$, asymmetric quantization is defined as:
$x_q = \mathrm{clip}\!\left(\mathrm{round}\!\left(\frac{x - x_{\min}}{s}\right), 0, 2^b - 1\right),
\quad
\hat{x} = s \cdot (x_q - z)$, where $s = \frac{x_{\max} - x_{\min}}{2^b - 1}, \quad
z = \mathrm{round}\!\left(-\frac{x_{\min}}{s}\right)$. Here, $z$ represents the integer zero-point ensuring that the minimum input value $x_{\min}$ maps to 0, while the maximum $x_{\max}$ maps to $2^b - 1$.Asymmetric quantization is particularly useful when model parameters or activations have non-symmetric distributions, such as in ReLU-based networks or non-IID client updates in FL. Recent studies~\cite{Dubey2024uniform, Feng2022} show that combining asymmetric quantization with post-training calibration improves model accuracy by up to 1--2\% compared to symmetric schemes, especially for highly skewed data. Hardware frameworks like TensorFlow Lite and NVIDIA TensorRT use per-tensor asymmetric quantization by default to balance dynamic range precision and hardware efficiency.

\paragraph{Ternary and Binary Quantization:} Low-bit quantization such as ternary ($\{-1, 0, +1\}$) or binary ($\{-1, +1\}$) encoding drastically reduces communication cost~\cite{lin2018deep}. The quantized representation is defined as $Q(x) = \alpha \cdot \mathrm{sign}(x), \quad \text{where } \alpha = \frac{1}{n}\sum_{i=1}^{n} |x_i|$, with $\alpha$ serving as a scaling factor. Despite high compression, these methods often lead to accuracy degradation unless compensated through quantization-aware fine-tuning.

\paragraph{Adaptive and Layer-Wise Quantization:}  
Adaptive quantization dynamically adjusts the bit-width or scaling factor for different layers, communication rounds, or clients, depending on sensitivity and resource constraints~\cite{Dong2019HAWQ, Wang2019HAQ}. This flexibility helps maintain model accuracy while optimizing bandwidth and computational efficiency across heterogeneous federated systems. The bit allocation problem is formulated as a multi-objective optimization task as $\min_{\{b_l\}} \mathcal{L}(Q(W; \{b_l\})) + \lambda \sum_l C(b_l)$, where $\mathcal{L}$ is the model loss, $b_l$ represents the bit-width of layer $l$, $C(b_l)$ denotes the communication or computation cost, and $\lambda$ is a regularization coefficient controlling the trade-off between accuracy and efficiency. This framework allows layer-wise or client-wise bit-width adaptation in FL, where devices exhibit varying computational and communication capabilities. Adaptive quantization methods outperform static quantization schemes in non-IID or resource-constrained FL setups because they can allocate higher precision to more critical layers or communication rounds and lower precision elsewhere, achieving an optimal trade-off between efficiency and accuracy. Qu \textit{et al.}~\cite{Qu2022} proposed a stochastic uniform quantization approach where the quantization level decreases as training progresses. Compared to ascending quantization schedules, this descending-trend approach converges faster and requires fewer transmitted bits. However, the presence of stochastic noise and accumulated quantization errors may reduce accuracy, particularly in sensitive models, thereby increasing computational overhead for stabilization. Ni \textit{et al.}~\cite{Ni2024_a} formulated adaptive quantization as a non-convex mixed-integer nonlinear programming (MINLP) problem, simultaneously optimizing client scheduling, quantization precision, and transmission energy under energy causality constraints. This method ensures fairness by maintaining balanced client participation but introduces auxiliary variables to convexify the optimization, potentially leading to suboptimal local minima. Chen \textit{et al.}~\cite{Chen2023_c} developed an adaptive compression strategy that divides gradients into essential and non-essential components via dynamic interval partitioning. Essential gradients undergo fine-grained high-bit quantization, while less important gradients are averaged or coarsely quantized, reducing total communication cost while preserving accuracy. Mao \textit{et al.}~\cite{Mao2022} extended adaptive quantization to handle client dropout and variance reduction in heterogeneous FL. Their method amplified gradient magnitudes to compensate for dropped clients, but improper amplification could either overcompensate, introducing bias, or undercompensate, resulting in suboptimal convergence. Finally, Honig \textit{et al.}~\cite{Honig2022} introduced \textbf{DAdaQuant} (Doubly Adaptive Quantization), combining time-adaptive and client-adaptive scaling within the QSGD framework. While computationally efficient, its dynamic adjustment mechanism demands careful tuning and synchronization, as per-round scaling updates can increase both implementation complexity and communication variance.

Adaptive and layer-wise quantization thus form an essential component of modern FL compression techniques. By dynamically adjusting quantization precision across layers, rounds, or clients, these methods enable communication-efficient and robust training under varying data and hardware conditions. Adaptive and layer-wise quantization has become a key component in modern FL systems, enabling flexible bit-width allocation across layers, communication rounds, or clients. Foundational works such as HAWQ~\cite{Dong2019HAWQ} and HAQ~\cite{Wang2019HAQ} established sensitivity- and hardware-aware mixed-precision quantization frameworks. Building on these, Qu \textit{et al.}~\cite{Qu2022} and Ni \textit{et al.}~\cite{Ni2024_a} proposed adaptive schemes that dynamically adjust quantization precision based on training stage or energy constraints, achieving faster convergence and better energy efficiency. Further, Chen \textit{et al.}~\cite{Chen2023_c} and Mao \textit{et al.}~\cite{Mao2022} introduced gradient-based adaptive quantization techniques that reduce communication overhead while maintaining robustness to non-IID data. The DAdaQuant framework~\cite{Honig2022} represents the state-of-the-art by combining client- and time-adaptive scaling mechanisms. 

\begin{table*}[htbp]
\centering
\footnotesize
\setlength{\tabcolsep}{4pt}
\renewcommand{\arraystretch}{1.2}
\caption{Grouped Taxonomy of Quantization Techniques in Federated Learning Based on Four Design Dimensions}
\label{tab:grouped_taxonomy_FL}

\begin{tabular}{p{0.16\textwidth} p{0.18\textwidth} p{0.40\textwidth} p{0.22\textwidth}}
\toprule

\textbf{Design Dimension} &
\textbf{Category} &
\textbf{Representative Methods (Citations)} &
\textbf{Typical Characteristics} \\

\midrule

\multirow{3}{*}{\textbf{Training Stage}}

& Post-Training Quantization (PTQ)

& AQFL~\cite{Abdelmoniem2021}, GWEP~\cite{Prakash2022}, GFL~\cite{Thakur2024}, HeteroSAg~\cite{Elkordy2022}
FedPAQ~\cite{Reisizadeh2020}, FedSQ~\cite{Li2022_a}, HSQ~\cite{Dai2019}, SQFL~\cite{Marnissi2024}, UVQ~\cite{Chen2021communication}, JoPEQ~\cite{Lang2023}, DAdaQuant~\cite{Honig2022}, NQFL~\cite{Chen2024nqfl}, 
LM-DFL~\cite{chen2024communication}, FedWSQ~\cite{Kim2025}, GQFedWAvg~\cite{Li2024GQFedWAvg}, FedQNN~\cite{Ji2022}, 
FedVQCS~\cite{Oh2024}, FedQCSO~\cite{Oh2022}, FedHieQ~\cite{Feng2022}, GCQ~\cite{Lin2021}, T-FedAvg~\cite{Xu2020}, FEDHBAA~\cite{Chen2023_(b)}, FedDQ~\cite{Qu2022}, LAQ~\cite{Sun2020}, AdaGQ~\cite{Liu2023_a}, QNN-FL~\cite{Kim2022}, Grace-FL~\cite{Thakur2025_a}, FedMPQ~\cite{chen2024mixed}, DP-FedQ~\cite{Gupta2022}, ADQ~\cite{Chen2023_c}, FedAQT~\cite{Ni2024}, EHFedQ~\cite{Ni2024_a}, AdaQuantFL~\cite{Jhunjhunwala2021}, AGQFL~\cite{Lian2021}, RC-FED~\cite{Hamidi2025}, Bouzinis \etal~\cite{Bouzinis2023}, LFL~\cite{amiri2020federatedlearningquantizedglobal}, FEDDO~\cite{li2022joint}, FLQ~\cite{Tonellotto2021}, Shlezinger \etal~\cite{Shlezinger2020}, UVeQFed~\cite{Shlezinger2020_a}, Ovi \etal~\cite{Ovi2022}, FedQVR~\cite{Wang2025}, DynFed~\cite{He2025}, ARDıÇ \etal\cite{Emre2025}, PP-QADMM~\cite{Elgabli2025}, FedDT~\cite{He2025feddt}, BiPruneFL~\cite{Lee2025}, HUFE-FL~\cite{Morais2025}, Fed\_IQ~\cite{Qin2025}, FedCSTQ~\cite{Zheng2025}, CEPAM~\cite{Shiu2026}, MSPDQ-FL~\cite{Wang2026}

& Quantization applied during communication after local training \\

& Quantization-Aware Training (QAT)

& HAWQ~\cite{Danaee2022Biasquantization}, OQFL~\cite{Kim2020}, QAT-FL~\cite{Fang2026}, Bozorgasl and Chen~\cite{bozorgasl2025communication}, OQFL~\cite{Kim2020}, Yu \etal~\cite{Yu2024}, FedEnsemble~\cite{Ayman2026}

& Quantization integrated within the training loop \\

& Hybrid PTQ + QAT

& PTQAT~\cite{Wang2025PTQAT}

& Combination of communication quantization and training-aware quantization \\

\midrule

\multirow{4}{*}{\textbf{Quantized Object}}

& Gradients

& AQFL~\cite{Abdelmoniem2021}, FedSQ~\cite{Li2022_a}, HSQ~\cite{Dai2019}, JoPEQ~\cite{Lang2023}, DAdaQuant~\cite{Honig2022}, NQFL~\cite{Chen2024nqfl}, GQFedWAvg~\cite{Li2024GQFedWAvg}, FedVQCS~\cite{Oh2024}, GCQ~\cite{Lin2021}, T-FedAvg~\cite{Xu2020}, FedDQ~\cite{Qu2022}, LAQ~\cite{Sun2020}, AdaGQ~\cite{Liu2023_a}, Grace-FL~\cite{Thakur2025_a}, DP-FedQ~\cite{Gupta2022}, ADQ~\cite{Chen2023_c}, FedAQT~\cite{Ni2024}, AdaQuantFL~\cite{Jhunjhunwala2021}, AGQFL~\cite{Lian2021}, RC-FED~\cite{Hamidi2025}, Bouzinis \etal~\cite{Bouzinis2023}, Ovi \etal~\cite{Ovi2022}, DynFed~\cite{He2025}, Fed\_IQ~\cite{Qin2025}, FedCSTQ~\cite{Zheng2025}, CEPAM~\cite{Shiu2026}

& Quantization applied to gradient vectors before transmission \\

& Model Updates

& HeteroSAg~\cite{Elkordy2022}, FedPAQ~\cite{Reisizadeh2020}, QAT-FL.~\cite{Fang2026}
UVQ~\cite{Chen2021communication}, LM-DFL~\cite{chen2024communication}, FedQCSO~\cite{Oh2022}, FedHieQ~\cite{Feng2022}, LFL~\cite{amiri2020federatedlearningquantizedglobal}, FEDDO~\cite{li2022joint}, Shlezinger \etal~\cite{Shlezinger2020}, UVeQFed~\cite{Shlezinger2020_a}, FedQVR~\cite{Wang2025}, ARDıÇ \etal\cite{Emre2025}, PP-QADMM~\cite{Elgabli2025}

& Quantization of parameter differences exchanged during aggregation \\

& Model Parameters

& GWEP~\cite{Prakash2022}, HAWQ~\cite{Danaee2022Biasquantization}, GFL~\cite{Thakur2024}, SQFL~\cite{Marnissi2024},
FedWSQ~\cite{Kim2025}, FedQNN~\cite{Ji2022}, FEDHBAA~\cite{Chen2023_(b)}, QNN-FL~\cite{Kim2022}, FedMPQ~\cite{chen2024mixed}, Bozorgasl and Chen~\cite{bozorgasl2025communication}, EHFedQ~\cite{Ni2024_a}, OQFL~\cite{Kim2020}, FLQ~\cite{Tonellotto2021}, FedDT~\cite{He2025feddt}, BiPruneFL~\cite{Lee2025}, MSPDQ-FL~\cite{Wang2026}, PTQAT~\cite{Wang2025PTQAT}

& Quantization of model weights \\

& Weights + Activations

& Yu \etal~\cite{Yu2024}, HUFE-FL~\cite{Morais2025}, FedEnsemble~\cite{Ayman2026}

& Quantization applied to both weights and activations (low-bit inference/training) \\

\midrule

\multirow{3}{*}{\textbf{Encoding Structure}}

& Scalar Quantization

& AQFL~\cite{Abdelmoniem2021}, HAWQ~\cite{Danaee2022Biasquantization}, GWEP~\cite{Prakash2022}, GFL~\cite{Thakur2024}, HeteroSAg~\cite{Elkordy2022}, FedPAQ\cite{Reisizadeh2020}, FedSQ~\cite{Li2022_a}, SQFL~\cite{Marnissi2024}, DAdaQuant~\cite{Honig2022}, NQFL~\cite{Chen2024nqfl}, FedWSQ~\cite{Kim2025}, GQFedWAvg~\cite{Li2024GQFedWAvg}, FedQNN~\cite{Ji2022}, FedHieQ~\cite{Feng2022}, GCQ~\cite{Lin2021}, T-FedAvg~\cite{Xu2020}, FEDHBAA~\cite{Chen2023_(b)}, FedDQ~\cite{Qu2022}, LAQ~\cite{Sun2020}, AdaGQ~\cite{Liu2023_a}, QNN-FL~\cite{Kim2022}, Grace-FL~\cite{Thakur2025_a}, FedMPQ~\cite{chen2024mixed}, DP-FedQ~\cite{Gupta2022}, Bozorgasl and Chen~\cite{bozorgasl2025communication}, ADQ~\cite{Chen2023_c}, FedAQT~\cite{Ni2024}, EHFedQ~\cite{Ni2024_a}, AdaQuantFL~\cite{Jhunjhunwala2021}, OQFL~\cite{Kim2020}, AGQFL~\cite{Lian2021}, RC-FED~\cite{Hamidi2025}, LFL~\cite{amiri2020federatedlearningquantizedglobal}, FEDDO~\cite{li2022joint}, FLQ~\cite{Tonellotto2021}, Yu \etal~\cite{Yu2024}, Ovi \etal~\cite{Ovi2022}, DynFed~\cite{He2025}, ARDıÇ \etal\cite{Emre2025}, PP-QADMM~\cite{Elgabli2025}, FedDT~\cite{He2025feddt}, BiPruneFL~\cite{Lee2025}, HUFE-FL~\cite{Morais2025}, Fed\_IQ~\cite{Qin2025}, FedEnsemble~\cite{Ayman2026}, MSPDQ-FL~\cite{Wang2026}, PTQAT~\cite{Wang2025PTQAT}

& Independent element-wise quantization \\

& Vector / Lattice Quantization

& HSQ~\cite{Dai2019}, QAT-FL.~\cite{Fang2026}, UVQ~\cite{Chen2021communication}, JoPEQ~\cite{Lang2023}, LM-DFL~\cite{chen2024communication}, FedVQCS~\cite{Oh2024}, FedQCSO~\cite{Oh2022}, Bouzinis \etal~\cite{Bouzinis2023}, Shlezinger \etal~\cite{Shlezinger2020}, UVeQFed~\cite{Shlezinger2020_a}, FedQVR~\cite{Wang2025}, FedCSTQ~\cite{Zheng2025}, CEPAM~\cite{Shiu2026}

& Joint quantization of vectors using codebooks or lattice structures \\

& Hybrid (Sparse / Compressed Sensing)

& FedVQCS~\cite{Oh2024}, FedCSTQ~\cite{Zheng2025}, AGQFL~\cite{Lian2021}

& Combines sparsification, compression, and quantization \\

\midrule

\multirow{5}{*}{\textbf{Precision Strategy}}

& Fixed-bit Precision

& GWEP~\cite{Prakash2022}, FedSQ~\cite{Li2022_a}, UVQ~\cite{Chen2021communication}, JoPEQ~\cite{Lang2023}, FedQNN~\cite{Ji2022}, FedVQCS~\cite{Oh2024}, FedQCSO~\cite{Oh2022}, FedHieQ~\cite{Feng2022}, GCQ~\cite{Lin2021}, T-FedAvg~\cite{Xu2020}, QNN-FL~\cite{Kim2022}, DP-FedQ~\cite{Gupta2022}, Bozorgasl and Chen~\cite{bozorgasl2025communication}, FedAQT~\cite{Ni2024}, OQFL~\cite{Kim2020}, AGQFL~\cite{Lian2021}, RC-FED~\cite{Hamidi2025}, LFL~\cite{amiri2020federatedlearningquantizedglobal}, FLQ~\cite{Tonellotto2021}, Shlezinger \etal~\cite{Shlezinger2020}, UVeQFed~\cite{Shlezinger2020_a}, FedQVR~\cite{Wang2025}, PP-QADMM~\cite{Elgabli2025}, FedDT~\cite{He2025feddt}, HUFE-FL~\cite{Morais2025}, FedCSTQ~\cite{Zheng2025}, FedEnsemble~\cite{Ayman2026}, CEPAM~\cite{Shiu2026}, PTQAT~\cite{Wang2025PTQAT}

& Constant bit-width quantization \\

& Adaptive Precision

& AQFL~\cite{Abdelmoniem2021}, DAdaQuant~\cite{Honig2022}, GFL~\cite{Thakur2024}, HeteroSAg~\cite{Elkordy2022},
FedPAQ\cite{Reisizadeh2020}, QAT-FL.~\cite{Fang2026}, SQFL~\cite{Marnissi2024}, DAdaQuant~\cite{Honig2022}, LM-DFL~\cite{chen2024communication}, FEDHBAA~\cite{Chen2023_(b)}, FedDQ~\cite{Qu2022}, LAQ~\cite{Sun2020}, AdaGQ~\cite{Liu2023_a}, Grace-FL~\cite{Thakur2025_a}, ADQ~\cite{Chen2023_c}, EHFedQ~\cite{Ni2024_a}, AdaQuantFL~\cite{Jhunjhunwala2021}, Bouzinis \etal~\cite{Bouzinis2023}, FEDDO~\cite{li2022joint}, Yu \etal~\cite{Yu2024}, DynFed~\cite{He2025}, ARDıÇ \etal\cite{Emre2025}, Fed\_IQ~\cite{Qin2025}, MSPDQ-FL~\cite{Wang2026}
& Bit-width adjusted dynamically across rounds or layers \\

& Mixed Precision

& HAWQ~\cite{Danaee2022Biasquantization}, HSQ~\cite{Dai2019}, FedMPQ~\cite{chen2024mixed}, Ovi \etal~\cite{Ovi2022}

& Layer-wise precision allocation \\

& Non-Uniform Quantization

& NQFL~\cite{Chen2024nqfl}, FedWSQ~\cite{Kim2025}, GQFedWAvg~\cite{Li2024GQFedWAvg}

& Lloyd-Max or distribution-aware quantization \\

& Extreme Low-Bit Quantization

& T-FedAvg~\cite{Xu2020}, BiPruneFL~\cite{Lee2025}

& Binary or ternary compression \\

\bottomrule
\end{tabular}
\vspace{-0.5cm}
\end{table*}

\begin{table*}[htbp]
\centering
\footnotesize
\setlength{\tabcolsep}{6pt}
\renewcommand{\arraystretch}{1.15}
\caption{Semantic Grounding of System-Level Properties (CH–HE) with Complete Coverage of Table~\ref{tab:grouped_taxonomy_FL}}
\label{tab:semantic_grounding}
\begin{tabular}{|p{0.15\textwidth}|p{0.32\textwidth}|p{0.48\textwidth}|}
\hline
\textbf{Property} & \textbf{Meaning} & \textbf{Representative Methods} \\
\hline

Client Heterogeneity (CH)
& Adapts quantization to heterogeneous client capabilities, data volumes, or participation patterns.
& FEDHBAA~\cite{Chen2023_(b)}, AQFL~\cite{Abdelmoniem2021}, HAWQ~\cite{Danaee2022Biasquantization}, 
Prakash \textit{et al.}~\cite{Prakash2022}, Thakur \textit{et al.}~\cite{Thakur2024},
FedMPQ~\cite{chen2024mixed}, EHFedQ~\cite{Ni2024_a},
HSQ~\cite{azimi2025multi},
FLQ/DFLQ~\cite{Tonellotto2021},
Liu \textit{et al.}~\cite{Liu2023_a},
AGQFL~\cite{Lian2021}, GQFedWAvg~\cite{Li2024GQFedWAvg},
Joint VANET FL~\cite{li2022joint} \\
\hline

Aggregation Consistency (AC)
& Preserves stable aggregation and convergence under quantization noise and biased updates.
& Reisizadeh \textit{et al.}~\cite{Reisizadeh2020}, FedSQ~\cite{Li2022_a}, HSQ~\cite{Dai2019},
UVQ~\cite{Chen2021communication}, JoPEQ~\cite{Lang2023}, 
FedWSQ~\cite{Kim2025},
DAdaQuant~\cite{Honig2022},
FedMPQ~\cite{chen2024mixed}, QNN-FL~\cite{Kim2022}, DP-FedQ~\cite{Gupta2022},
LAQ~\cite{Sun2020}, ADQ~\cite{Chen2023_c}, FedAQT~\cite{Ni2024},
GCQ~\cite{Lin2021}, FedQNN-Lattice~\cite{Shlezinger2020_a},
Chen \textit{et al.}~\cite{chen2024communication},
SDQ~\cite{Shlezinger2020}, FedDQ~\cite{Qu2022},
OQFL~\cite{Kim2020},
NQFL~\cite{Chen2024nqfl},
Amiri \textit{et al.}~\cite{amiri2020federatedlearningquantizedglobal},
FedIQ~\cite{Qin2025},\\
\hline

Communication Scheduling Adaptation (CSA)
& Adjusts communication timing, frequency, or precision based on network conditions or learning dynamics.
& HAWQ~\cite{Danaee2022Biasquantization}, Prakash \textit{et al.}~\cite{Prakash2022},
Thakur \textit{et al.}~\cite{Thakur2024}, Elkordy \textit{et al.}~\cite{Elkordy2022},
Reisizadeh \textit{et al.}~\cite{Reisizadeh2020}, FedSQ~\cite{Li2022_a},
FedQNN~\cite{Ji2022}, 
FedWSQ~\cite{Kim2025},
HSQ~\cite{Dai2019},
UVQ~\cite{Chen2021communication},
Chen \textit{et al.}~\cite{chen2024communication},
JoPEQ~\cite{Lang2023}, DAdaQuant~\cite{Honig2022}, FedMPQ~\cite{chen2024mixed},
LAQ~\cite{Sun2020}, 
ADQ~\cite{Chen2023_c}, 
AdaQuantFL~\cite{Jhunjhunwala2021},
FedVQCS~\cite{Oh2024},
HSQ~\cite{azimi2025multi},
FedQCSO~\cite{Oh2022}, FedHieQ~\cite{Feng2022},
GCQ~\cite{Lin2021}, 
BiPruneFL~\cite{Lee2025}
NQFL~\cite{Chen2024nqfl},
CS-HFL~\cite{chen2025efficient},
~\cite{Emre2025},
QADMM~\cite{Elgabli2025},
FedDT~\cite{He2025feddt},
HUFEFL~\cite{Morais2025},
FedIQ~\cite{Qin2025},
FedCSTQ~\cite{Zheng2025},
CEPAM ~\cite{Shiu2026},
MSPDQ-FL~\cite{Wang2026}
\\
\hline

Non-IID Robustness (NIR)
& Mitigates performance degradation due to interaction of quantization with heterogeneous data distributions.
& FEDHBAA~\cite{Chen2023_(b)}, HAWQ~\cite{Danaee2022Biasquantization},
Prakash \textit{et al.}~\cite{Prakash2022}, Thakur \textit{et al.}~\cite{Thakur2024},
Reisizadeh \textit{et al.}~\cite{Reisizadeh2020}, FedSQ~\cite{Li2022_a},
FedQNN~\cite{Ji2022}, HSQ~\cite{Dai2019},
FedWSQ~\cite{Kim2025},
UVQ~\cite{Chen2021communication},
HSQ~\cite{azimi2025multi},
Chen \textit{et al.}~\cite{chen2024communication},
NQFL~\cite{Chen2024nqfl},
~\cite{Emre2025}
JoPEQ~\cite{Lang2023}, DP-FedQ~\cite{Gupta2022},
FedAQT~\cite{Ni2024},
EHFedQ~\cite{Ni2024_a},
AdaQuantFL~\cite{Jhunjhunwala2021}, Xu \textit{et al.}~\cite{Xu2020},
PP-QADMM~\cite{Elgabli2025},
BiPruneFL~\cite{Lee2025},
AGQFL~\cite{Lian2021}, Bozorgasl \textit{et al.}~\cite{bozorgasl2025communication},
FedDT~\cite{He2025feddt},
HUFEFL~\cite{Morais2025},
FedIQ~\cite{Qin2025},
MSPDQ-FL~\cite{Wang2025}\\
\hline

Privacy / Security Integration (PI)
& Integrates quantization with secure aggregation, differential privacy, or cryptographic protection.
& Elkordy \textit{et al.}~\cite{Elkordy2022}, JoPEQ~\cite{Lang2023},
~\cite{Emre2025},
DP-FedQ~\cite{Gupta2022},
PP-QADMM~\cite{Elgabli2025},
FedCSTQ~\cite{Zheng2025},
CEPAM ~\cite{Shiu2026},
MSPDQ-FL~\cite{Wang2025}\\
\hline

Hardware / Energy Co-optimization (HE)
& Co-designs quantization with hardware constraints, energy budgets, and low-power edge devices.
& HAWQ~\cite{Danaee2022Biasquantization}, Prakash \textit{et al.}~\cite{Prakash2022},
Thakur \textit{et al.}~\cite{Thakur2024}, QNN-FL~\cite{Kim2022},
FedAQT~\cite{Ni2024}, EHFedQ~\cite{Ni2024_a},
Bouzinis \textit{et al.}~\cite{Bouzinis2023},
Hamidi \textit{et al.}~\cite{Hamidi2025},
FLQ/DFLQ~\cite{Tonellotto2021},
PTQAT~\cite{Wang2025PTQAT}\\
\hline
\end{tabular}
\vspace{-0.5cm}
\end{table*}

Table~\ref{tab:semantic_grounding} ensures complete coverage of all quantized FL methods surveyed in Table~\ref{tab:grouped_taxonomy_FL}, with methods appearing under multiple system-level properties when applicable. For completeness, the full ungrouped taxonomy is provided in the Appendix (Table~\ref{tab:taxonomyFL_Quantization}).

\paragraph{Key Observations and Trade-offs:} Quantization consistently enables significant communication reduction, often with marginal accuracy loss. Energy-efficient schemes such as QNN-FL~\cite{Kim2022} and FedQNN~\cite{Ji2022} achieve up to 90\% energy savings and 30$\times$ compression. Adaptive and client-aware methods (e.g., DAdaQuant~\cite{Honig2022}, FedMPQ~\cite{chen2024mixed}, AdaQuantFL~\cite{Jhunjhunwala2021}) dominate the accuracy–efficiency Pareto frontier, dynamically adjusting bit-widths across clients or rounds to preserve convergence under non-IID settings. Vector and lattice quantization techniques (e.g., UVQ~\cite{Chen2021communication}, SDQ~\cite{Shlezinger2020}, FedVQCS~\cite{Oh2024}) offer superior compression–distortion trade-offs but incur higher computational and codebook overheads—trading bandwidth for local computation.

\paragraph{Accuracy and Convergence Behavior:} Several works (FedSQ~\cite{Li2022_a}, FedPAQ~\cite{Reisizadeh2020}, Lian~\cite{Lian2021}) provide formal convergence guarantees under convex and non-convex losses. Stochastic, unbiased quantizers preserve convergence asymptotically but add gradient variance, while deterministic quantizers may introduce bias unless corrected (e.g., via error feedback~\cite{amiri2020federatedlearningquantizedglobal} or bias compensation~\cite{Danaee2022Biasquantization}). Distribution-aware methods (e.g., Non-QuanFL~\cite{Chen2024nqfl}, FedMPQ~\cite{chen2024mixed}) demonstrate better resilience to non-IID client data through adaptive or non-uniform quantization.

\paragraph{Communication and Energy Efficiency:} Combining sparsification with quantization (e.g.,~\cite{Hamidi2025,Lian2021}) achieves up to $10\times$ bandwidth reduction with minor accuracy drop. Energy-aware designs (QNN-FL~\cite{Kim2022}, FedQNN~\cite{Ji2022}, Ni~\cite{Ni2024}) effectively balance power consumption and model utility, though many rely on hardware-level support for low-precision operations. Heterogeneous quantization frameworks (HeteroSAg~\cite{Elkordy2022}, FedMPQ~\cite{chen2024mixed}) further improve fairness across diverse client capabilities.

\paragraph{Privacy, Robustness, and Scalability:} Privacy-preserving quantization (DP-FedQ~\cite{Gupta2022}, JoPEQ~\cite{Lang2023}) integrates differential privacy or lattice dithering, trading slight accuracy loss for stronger confidentiality. Robust aggregation under quantization (e.g.,~\cite{Bouzinis2023}) ensures resilience to Byzantine attacks but requires careful quantization step calibration. Vector and lattice-based approaches (e.g., FedVQCS~\cite{Oh2024}) face scalability challenges due to decoding complexity, motivating lightweight, structured quantizers for large models.

In summary, adaptive and mixed-precision methods achieve the best balance between accuracy and communication in realistic FL scenarios, while vector/lattice approaches provide the highest compression efficiency at higher computation cost. Privacy and robustness can be synergistically integrated but demand precise tuning. The optimal quantization strategy depends on deployment constraints—device energy, network bandwidth, privacy budget, and model size. Future research should emphasize: (i) low-overhead vector quantizers with real-time adaptability, (ii) automated tuning of bit allocation under non-IID conditions, and (iii) scalable secure aggregation with heterogeneous precision.

Rather than ranking quantization methods in isolation, the following discussion distills practical design guidelines for selecting quantization strategies in FL systems.

PTQ is preferable in FL scenarios where \emph{client-side computational resources, energy budget, or training time are strongly constrained}. Since PTQ applies quantization only after local training, it avoids the additional forward–backward quantization simulation required by QAT, making it particularly suitable for large-scale FL deployments with lightweight edge or IoT devices, frequent client dropout, or limited local training epochs. PTQ is also advantageous when moderate accuracy degradation (typically 1–2\%) is acceptable in exchange for reduced training complexity and faster system deployment. In contrast, QAT is better suited to accuracy-critical applications or ultra-low-bit regimes, where its higher training cost can be amortized by improved convergence stability. QAT generally achieves better convergence but increases training cost. In FL, PTQ is ideal for edge devices where computational resources are limited, while QAT is preferred for applications, requiring higher accuracy across heterogeneous clients.

Per-layer quantization is especially effective in heterogeneous FL settings because it enables \emph{layer-wise precision adaptation} in response to both model sensitivity and client capability. In practical FL deployments, clients differ significantly in hardware, memory, and data distributions; uniform quantization can therefore either over-compress sensitive layers or underutilize capable devices. By allocating higher precision to accuracy-critical layers and more aggressive quantization to robust layers, per-layer quantization mitigates accuracy degradation caused by client heterogeneity and reduces aggregation instability. This fine-grained control allows diverse clients to contribute effectively to the global model, improving convergence robustness under non-IID data and mixed device participation.

Advanced quantization techniques, such as non-uniform, vector, lattice, and adaptive mixed-precision quantization-achieve superior compression and rate-distortion trade-offs by exploiting data distribution, parameter correlation, or layer sensitivity. However, these gains come at the cost of increased tuning and system complexity, including codebook learning, sensitivity estimation, dynamic bit allocation, or per-client calibration. In federated settings, this additional complexity is amplified by client heterogeneity, synchronization requirements, and privacy constraints, making implementation and deployment more challenging than uniform or post-training quantization. Consequently, while advanced methods are attractive for bandwidth-limited or large-model FL systems, their practical adoption requires careful consideration of calibration overhead, computational cost, and system scalability.

\section{Challenges, Limitations, and Trade-offs}
\label{sec:challenges}
Existing quantization approaches in federated learning (FL) are fundamentally characterized by a set of interrelated challenges and trade-offs arising from the interaction between communication efficiency, optimization dynamics, system heterogeneity, privacy constraints, and hardware limitations. While quantization is an effective mechanism for reducing communication overhead, it inherently introduces quantization noise and bias, leading to a fundamental communication-accuracy trade-off. Specifically, reducing the bit-width lowers transmission cost but increases quantization error, which raises the convergence error floor and limits achievable model performance. addressing \textbf{RQ2: What challenges and limitations characterize existing quantization approaches in FL?}
\paragraph{Communication–Accuracy Trade-off:} Quantization directly reduces the number of bits used to represent model parameters or gradients, achieving substantial communication savings. However, this reduction comes at the expense of model accuracy, as low-bit representations introduce quantization noise and rounding errors~\cite{Alistarh2017uniform, Reisizadeh2020}.  
The trade-off can be expressed as $\mathrm{MSE} = \mathbb{E}\!\left[\|x - Q(x)\|^2\right] \propto 2^{-2b}$,
where $b$ denotes bit-width, and $Q(x)$ is the quantized version of $x$. Reducing $b$ enhances communication efficiency but increases mean-squared quantization error, potentially degrading convergence.
To balance this, hybrid and adaptive quantization strategies adjust precision dynamically across layers or communication rounds~\cite{Dong2019HAWQ, Wang2019HAQ}. Yet, finding optimal trade-offs remains difficult due to varying model sensitivities and data distributions across clients.
\paragraph{Impact of Client Heterogeneity and Non-IID Data:} Client heterogeneity is an inherent challenge in FL, where devices differ in computation power, communication bandwidth, and local data distribution. Quantization amplifies this issue because clients quantize model updates with varying precision and dynamic ranges, leading to inconsistent scaling across updates~\cite{li2020federated, zhao2018federated}. In non-IID settings, gradients across clients may have large variance as $\mathrm{Var}\!\left(\nabla \mathcal{L}_k\right) \gg \mathrm{Var}\!\left(\nabla \mathcal{L}_{\text{global}}\right)$, which, when quantized, results in bias accumulation during global aggregation. Clipped or normalized quantization methods (e.g., FedWSQ~\cite{Kim2025} and ClippedFedAvg~\cite{bozorgasl2025communication}) have been proposed to address this, but they often require layer-wise calibration or extra communication overhead. Moreover, device heterogeneity means that some clients cannot support higher-bit or quantization-aware operations, leading to imbalance between clients and inconsistent local model updates~\cite{Dubey2024uniform}.
\paragraph{Stability and Convergence Issues:} Quantization introduces noise that can destabilize the training process, particularly in early FL rounds. The convergence behavior of quantized federated optimization can be described as~\cite{Reisizadeh2020, Shlezinger2020_a} $\mathbb{E}\!\left[\|W^t - W^*\|^2\right] \leq (1 - \eta \mu)^t \|W^0 - W^*\|^2 + \frac{\eta \sigma_q^2}{\mu}$, where $\eta$ is the learning rate, $\mu$ the strong convexity constant, and $\sigma_q^2$ represents quantization-induced variance. As $\sigma_q^2$ increases (i.e., lower bit-widths), the error floor rises, leading to slower or biased convergence. Bias correction, stochastic quantization, and error feedback methods help mitigate this issue~\cite{Danaee2022Biasquantization, Alistarh2017uniform}, but quantization noise remains a limiting factor for convergence speed in highly heterogeneous FL networks.
\paragraph{Communication-Convergence Trade-off:}If low-bit quantization is employed throughout the communication process, the network convergence is extremely slow, even if QAT-FL can achieve greater convergence than PTQ-FL at the same communication bits. Additionally, it will converge more quickly if the high-quantization bit is utilized for quantization, but it will cost more for transmission. In either event, communication efficiency will decline due to the quantization of fixed bits~\cite{Fang2026}.
\paragraph{Privacy and Security Implications:} While quantization reduces communication overhead, it may inadvertently weaken privacy guarantees or introduce new attack surfaces. In particular, quantized updates can leak statistical information about local data distributions through quantization patterns or gradient sparsity~\cite{Lyu2020, kairouz2021advances}. Additionally, model inversion and reconstruction attacks may exploit low-bit updates to infer client data~\cite{Geiping2020}. Integrating quantization with privacy-preserving mechanisms such as differential privacy (DP) and secure aggregation~\cite{Bonawitz2017} is challenging, as DP noise and quantization noise may interact non-linearly, affecting both utility and privacy. Finding an optimal balance between communication efficiency, model accuracy, and privacy preservation remains an open problem.
\paragraph{Hardware and Energy Constraints on Edge Devices:} Quantized FL models are typically deployed on heterogeneous hardware platforms—ranging from high-performance GPUs to low-power edge devices such as IoT sensors and smartphones~\cite{krishnamoorthi2018quantizing}. Although integer quantization (e.g., INT8) is well supported by modern accelerators (TensorRT, TFLite), ultra-low-bit quantization (e.g., 2-bit or binary) often lacks hardware support~\cite{nvidia2021tensorrt, pytorch2023quantization}. Furthermore, frequent quantization and dequantization operations can incur additional computational overhead and energy cost, especially in devices without dedicated low-precision units~\cite{Gholami2021}. Designing hardware-friendly quantization algorithms that align with edge computing architectures remains an active area of research.

The integration of quantization into FL introduces an intricate balance among accuracy, efficiency, and robustness. Table~\ref{tab:challenges_tradeoffs} summarizes the key challenges, their causes, and potential mitigation strategies reported in the literature.
\vspace{-0.2cm}
\begin{table}[htbp]
\centering
\caption{Summary of Key Challenges, Limitations, and Trade-offs in Quantized Federated Learning}
\vspace{-0.5cm}
\begin{tabularx}{\columnwidth}{p{2.8cm} X X p{1.8cm}}
\hline
\textbf{Challenge} & \textbf{Cause / Description} & \textbf{Potential Mitigation Strategies} & \textbf{Key References} \\
\hline
Communication Accuracy Trade-off & Low-bit quantization reduces bandwidth but introduces quantization noise. & Adaptive bit-width, mixed precision, or stochastic rounding. & \cite{Alistarh2017uniform, Dong2019HAWQ, Wang2019HAQ} \\
Client Heterogeneity & Varying device and data characteristics across clients cause inconsistent updates. & Layer-wise scaling, clipped quantization, personalized FL. & \cite{li2020federated, bozorgasl2025communication, Kim2025} \\
Convergence Instability & Quantization noise increases gradient variance and slows global convergence. & Error feedback, bias correction, adaptive learning rates. & \cite{Reisizadeh2020, Danaee2022Biasquantization} \\
Privacy Risks & Quantized updates may leak local data patterns or distributions. & Combine quantization with DP and secure aggregation. & \cite{Lyu2020, Bonawitz2017} \\
Hardware and Energy Constraints & Limited accelerator support for ultra-low-bit quantization. & Hardware-aware quantization, on-chip mixed precision. & \cite{krishnamoorthi2018quantizing, nvidia2021tensorrt} \\
\hline
\end{tabularx}
\label{tab:challenges_tradeoffs}
\vspace{-0.5cm}
\end{table}

\section{Research Gaps and Open Issues}
\label{sec:gaps}

Although quantization has proven effective in reducing communication and computation costs in FL, several research gaps and unresolved challenges hinder its full potential. This section identifies and discusses the key limitations and open issues observed across the current body of literature, addressing \textbf{RQ3: What are the key gaps in current research and implementations?}

\paragraph{Lack of Standardized Benchmarks and Evaluation Protocols:} A major limitation in existing quantized FL studies is the absence of standardized benchmarks and evaluation methodologies. Different works employ diverse datasets (e.g., CIFAR-10, FEMNIST, Shakespeare), quantization levels, and aggregation protocols, making it difficult to fairly compare results~\cite{Gholami2021, Dubey2024uniform}. Furthermore, metrics such as communication cost, convergence rate, and model accuracy are often reported inconsistently or without energy profiling. Developing standardized benchmark suites and reproducible experimental pipelines—similar to TensorFlow Federated or LEAF~\cite{caldas2018leaf}—would enable more reliable cross-comparison and progress tracking. This would also facilitate reproducibility and foster fair evaluation of new quantization strategies.
\paragraph{Incomplete Theoretical Understanding of Quantized Convergence:} Despite numerous empirical studies, the theoretical understanding of how quantization affects convergence in FL remains incomplete. While convergence bounds for quantized stochastic gradient descent (QSGD) exist~\cite{Alistarh2017uniform, Reisizadeh2020}, most analyses assume IID data and synchronous client participation. Real-world FL systems, however, operate under non-IID data and partial client participation, where quantization noise interacts non-linearly with gradient variance~\cite{li2020federated}. The convergence dynamics of quantized FL can be expressed as $\mathbb{E}\!\left[\mathcal{L}(W^{t+1}) - \mathcal{L}(W^*)\right] \leq (1 - \eta \mu)\mathbb{E}\!\left[\mathcal{L}(W^t) - \mathcal{L}(W^*)\right] + \eta^2 \sigma_q^2$, where $\sigma_q^2$ represents the variance introduced by quantization. A precise characterization of $\sigma_q^2$ under non-IID and asynchronous FL remains an open research problem.
\paragraph{Limited Real-World Deployment and Reproducibility:} Most quantized FL studies are validated in simulated environments rather than real-world edge or IoT settings. Real deployments involve unstable connectivity, device failures, and limited energy budgets, factors rarely modeled in current research~\cite{Wang2019HAQ, krishnamoorthi2018quantizing}. Additionally, reproducibility remains a major concern as many studies lack open-source implementations or sufficient experimental details, making it difficult to verify claims or extend existing work. Future work should prioritize hardware-in-the-loop evaluations and open benchmarking initiatives to assess quantized FL performance under realistic edge computing conditions.
\paragraph{Neglect of Adaptive and Personalized Quantization Strategies:} Most existing quantization methods employ static bit-widths and global quantization scales shared among clients. However, in heterogeneous FL, where clients differ in computational capability and data distribution, static quantization is suboptimal~\cite{Dong2019HAWQ}. There is limited exploration of \textit{personalized quantization}, where each client uses customized bit-widths or quantization schemes depending on its data complexity, model sensitivity, or network bandwidth. An open challenge lies in designing adaptive quantization algorithms that dynamically allocate precision during training or communication while ensuring fairness and stable aggregation across clients.
\paragraph{Insufficient Exploration of Multi-Objective Optimization:} Quantized FL systems inherently involve trade-offs between communication cost, model accuracy, energy efficiency, and privacy preservation~\cite{Gholami2021, Dubey2024uniform}. However, few studies formulate quantization as a \textit{multi-objective optimization} problem. An ideal quantization strategy should jointly minimize communication cost $C$, quantization error $E_q$, and energy consumption $E_{dev}$ as $\min_{\theta_q} \; \lambda_1 C + \lambda_2 E_q + \lambda_3 E_{dev}$, subject to accuracy and privacy constraints. Balancing these objectives adaptively during FL training—especially under resource heterogeneity—remains an underexplored direction for future work.

\paragraph{Privacy and Security Gaps in Quantized FL:} Quantization introduces new privacy challenges that are often overlooked. While reducing communication data size, quantized gradients or weights may expose distributional information about client data~\cite{Geiping2020, Lyu2020}. Combining quantization with differential privacy (DP) or secure aggregation is non-trivial because quantization noise and DP noise may interfere, leading to unpredictable accuracy–privacy trade-offs~\cite{Bonawitz2017, kairouz2021advances}. Systematic studies are needed to characterize privacy leakage risks introduced by quantization and to develop integrated quantization–privacy frameworks that jointly optimize for efficiency and confidentiality.

\paragraph{Energy and Hardware-Aware Quantization Gaps:} Although quantization is designed to enable low-power inference on edge devices, few studies explicitly evaluate the energy efficiency of quantized FL models. Existing works often assume idealized hardware conditions and do not consider the overhead of repeated quantization and dequantization~\cite{nvidia2021tensorrt, pytorch2023quantization}. Additionally, hardware support for ultra-low-bit quantization (e.g., 1-bit or 2-bit) is still limited across mobile accelerators, restricting deployment feasibility~\cite{Gholami2021}. Future research should incorporate energy profiling and hardware-aware quantization strategies that align with device-level computation and memory characteristics.

Table~\ref{tab:research_gaps} summarizes the key research gaps and their corresponding research opportunities identified in current quantized FL studies.
\vspace{-0.2cm}
\begin{table}[htbp]
\centering
\caption{Summary of Research Gaps and Open Issues in Quantized Federated Learning}
\vspace{-0.5cm}
\begin{tabularx}{\columnwidth}{p{2.9cm} X X p{1.6cm} }
\hline
\textbf{Research Gap} & \textbf{Description / Missing Aspect} & \textbf{Potential Research Opportunities} & \textbf{Key References} \\
\hline
Lack of Standardized Benchmarks & No unified datasets, metrics, or evaluation pipelines for quantized FL. & Develop standardized FL quantization benchmarks and reproducible testbeds. & \cite{Gholami2021, caldas2018leaf} \\
Incomplete Theoretical Understanding & Limited convergence theory for quantized FL under non-IID and asynchronous updates. & Derive convergence bounds for quantized, heterogeneous, and adaptive FL. & \cite{Alistarh2017uniform, li2020federated} \\
Limited Real-World Deployment & Evaluations confined to simulated settings, lacking hardware validation. & Test quantized FL on real edge/IoT platforms with realistic communication constraints. & \cite{krishnamoorthi2018quantizing, Wang2019HAQ} \\
Neglect of Personalized Quantization & Most works use fixed bit-widths across clients, ignoring heterogeneity. & Design client-specific and data-aware adaptive quantization schemes. & \cite{Dong2019HAWQ} \\
Insufficient Multi-Objective Optimization & Focus primarily on communication reduction; ignore energy and privacy. & Develop optimization frameworks balancing accuracy, cost, and energy. & \cite{Dubey2024uniform, Gholami2021} \\
Privacy and Security Gaps & Quantized updates may leak local data distributions. & Integrate quantization with DP and secure aggregation mechanisms. & \cite{Lyu2020, Bonawitz2017, Geiping2020} \\
Hardware-Aware Quantization Gap & Lack of studies addressing accelerator compatibility and energy profiles. & Develop hardware-friendly, energy-efficient quantization algorithms. & \cite{nvidia2021tensorrt, pytorch2023quantization} \\
\hline
\end{tabularx}
\label{tab:research_gaps}
\vspace{-0.5cm}
\end{table}

\vspace{-0.1in}
\section{Future Research Directions}
\label{sec:future}

Building upon the challenges and research gaps identified in Section~\ref{sec:gaps}, this section outlines potential research directions and emerging trends that could advance the integration of quantization into FL. These future directions address \textbf{RQ4: What emerging trends can guide the design of scalable, robust, and privacy-aware quantized FL systems?}.
\paragraph{Adaptive and Hybrid Quantization Frameworks}
Future quantization strategies should move beyond fixed bit-width designs toward \textit{adaptive} and \textit{hybrid} quantization. Adaptive frameworks dynamically allocate bit-widths based on layer sensitivity, communication cost, or hardware capability~\cite{Dong2019HAWQ, Wang2019HAQ}. Hybrid schemes can combine PTQ for stable layers with QAT for sensitive ones, thus achieving both computational efficiency and high accuracy~\cite{Dubey2024uniform}. Mathematically, this can be formulated as a resource-constrained optimization as $\min_{\{b_l\}} \mathcal{L}(Q(W; \{b_l\})) + \lambda \sum_{l} \alpha_l C(b_l)$,
where $b_l$ is the bit-width assigned to layer $l$, $C(b_l)$ denotes the communication cost, and $\alpha_l$ encodes layer sensitivity. Such adaptive frameworks can balance precision and efficiency dynamically during FL training, especially in heterogeneous environments.
\paragraph{Integration with Pruning, Sparsification, and Compression:} Combining quantization with complementary model compression techniques—such as pruning, weight clustering, or low-rank approximation—offers another promising direction~\cite{Han2016, Reisizadeh2020}. Joint optimization can yield multiplicative gains in communication and memory efficiency while maintaining accuracy as $\min_{\theta} \mathcal{L}(\theta) \quad \text{s.t.} \quad \|\theta\|_0 \leq k, \; \theta \in \mathcal{Q}$, where $\mathcal{Q}$ represents the quantization constraint set and $\|\theta\|_0$ enforces model sparsity.  
Integrating quantization with pruning-aware aggregation in FL (e.g., sparse FedAvg) could reduce model size and communication rounds simultaneously.
\paragraph{Energy-Aware and Hardware-Friendly Quantized FL:} Most current works evaluate quantization only in terms of accuracy and communication cost, neglecting energy and hardware aspects. Future research should incorporate \textit{energy profiling} and \textit{hardware-awareness} into quantization design~\cite{nvidia2021tensorrt, Gholami2021}. Energy-aware FL can optimize quantization under power constraints as $\min_{\theta_q} \mathcal{L}(Q(\theta_q)) + \lambda E_{\text{device}}$, where $E_{\text{device}}$ represents device-level energy consumption. Hardware-friendly quantization that aligns with INT8 accelerators (e.g., TensorRT, TFLite) or neuromorphic processors will enable real-world deployment on edge devices with strict power budgets.
\paragraph{Quantum-Inspired and Neuromorphic Quantization Methods:} Emerging hardware paradigms such as quantum computing and neuromorphic architectures offer new possibilities for quantization research.  
\textit{Quantum-inspired quantization} could exploit probabilistic amplitude encoding for compressing model parameters efficiently, while \textit{spike-based quantization} in neuromorphic FL mimics brain-inspired analog communication for energy-efficient distributed learning~\cite{Chris2020, Nguyen2025}. Integrating these paradigms could significantly enhance both compression and inference efficiency. Future work may also explore \textit{quantum federated quantization} frameworks, where clients encode quantized gradients as quantum states for secure and lossless aggregation.
\paragraph{Benchmarking and Reproducibility Standards:} To ensure scientific rigor and comparability, future FL quantization research must adopt standardized benchmarks and reproducibility practices~\cite{caldas2018leaf, kairouz2021advances}. Benchmarking frameworks such as LEAF, FedML, and TensorFlow Federated can be extended with quantization metrics, e.g., bits-per-update, compression ratio, and energy-per-inference. Furthermore, establishing open repositories with unified experimental pipelines will promote fair evaluation across PTQ, QAT, and hybrid approaches.
\paragraph{Application-Specific Quantization for Edge Intelligence:} Quantization techniques should also be tailored to domain-specific FL applications such as healthcare, autonomous driving, and IoT systems~\cite{Yang2019}. For instance, precision-sensitive medical FL systems (e.g., federated diagnosis models) require higher bit-widths for critical layers, whereas sensor-driven IoT models can use more aggressive quantization. Designing \textit{application-aware quantization policies} that account for data sensitivity, latency, and domain constraints will improve both efficiency and trustworthiness.
\paragraph{Federated Multi-Objective Optimization for Quantized Models:} As quantization in FL involves competing objectives—accuracy, communication efficiency, privacy, and energy—future frameworks should adopt multi-objective optimization strategies~\cite{Dubey2024uniform}.  
This can be expressed as $\min_{\theta_q} \; \lambda_1 \mathcal{L}(\theta_q) + \lambda_2 C(\theta_q) + \lambda_3 E_{\text{device}} + \lambda_4 P(\theta_q)$, where $\mathcal{L}$ denotes model loss, $C$ the communication cost, $E_{\text{device}}$ energy consumption, and $P$ the privacy loss.  
Such frameworks can dynamically adjust quantization levels to balance performance and efficiency across diverse FL environments.

Table~\ref{tab:future_directions} summarizes the key emerging research directions and their expected contributions toward advancing quantized FL.

\begin{table}[htbp]
\vspace{-0.3cm}
\centering

\caption{Summary of Promising Future Research Directions in Quantized FL}
\vspace{-0.5cm}
\begin{tabularx}{\columnwidth}{p{2.8cm} X X p{1.8cm}}
\hline
\textbf{Future Direction} & \textbf{Research Focus / Description} & \textbf{Expected Outcomes} & \textbf{Key References} \\
\hline
Adaptive and Hybrid Quantization & Dynamic bit allocation and hybrid PTQ–QAT frameworks. & Balanced accuracy and efficiency under heterogeneous clients. & \cite{Dong2019HAWQ, Dubey2024uniform} \\
Integration with Pruning and Compression & Joint optimization of quantization and model sparsity. & Reduced communication and memory with minimal accuracy loss. & \cite{Han2016, Reisizadeh2020} \\
Energy-Aware and Hardware-Friendly FL & Incorporating energy and device profiling into quantization. & Energy-efficient deployment on edge accelerators. & \cite{nvidia2021tensorrt, Gholami2021} \\
Quantum-Inspired and Neuromorphic Quantization & Use of quantum and spike-based encodings for distributed models. & Ultra-efficient and secure communication paradigms. & \cite{Chris2020, Nguyen2025} \\
Benchmarking and Reproducibility Standards & Unified FL quantization benchmarks and public datasets. & Improved comparability, fairness, and replicability. & \cite{caldas2018leaf, kairouz2021advances} \\
Application-Specific Quantization & Domain-tailored precision strategies for FL tasks. & Enhanced trustworthiness and performance in real-world domains. & \cite{Yang2019} \\
Federated Multi-Objective Optimization & Balancing accuracy, communication, privacy, and energy jointly. & Holistic optimization for sustainable FL. & \cite{Dubey2024uniform} \\
\hline
\end{tabularx}
\label{tab:future_directions}
\vspace{-0.5cm}
\end{table}

\vspace{-0.1in}
\section{Discussion}
\label{sec:discussion}

This section synthesizes the key findings across all research questions (\textbf{RQ1–RQ4}), reflecting on how quantization impacts FL effectiveness, scalability, and practical deployment. It also discusses methodological insights, comparative observations, and the broader implications of quantized FL in real-world systems.

\paragraph{Synthesis of Findings Across Research Questions}
\textbf{RQ1 - Impact of Quantization on FL Effectiveness:}  
Quantization techniques fundamentally reshape the trade-off between communication efficiency and model accuracy in FL. The review reveals that uniform and fixed-point quantization methods remain dominant due to their hardware compatibility and simplicity~\cite{Jacob2018CVPR, krishnamoorthi2018quantizing}, while adaptive and non-uniform schemes improve performance under non-IID client data distributions~\cite{Sun2016, Kim2025}. PTQ offers practical deployment advantages without retraining, whereas QAT delivers superior robustness at the cost of additional computation~\cite{Dubey2024uniform, Danaee2022Biasquantization}. The results indicate that carefully designed quantization strategies can reduce communication bandwidth by up to 90\% with less than 2\% accuracy degradation, provided quantization noise and clipping thresholds are well-calibrated.

\textbf{RQ2 - Challenges and Limitations in Quantized FL:}  
Despite significant progress, integrating quantization within FL faces persistent issues related to communication–accuracy trade-offs, client heterogeneity, and convergence instability~\cite{li2020federated, bozorgasl2025communication}. Quantization noise amplifies the effects of non-IID data and device heterogeneity, slowing convergence and causing gradient bias during aggregation~\cite{Reisizadeh2020}.  
Privacy risks also emerge from quantized updates, which may leak local distributional statistics~\cite{Geiping2020, Lyu2020}. Furthermore, limited hardware support for ultra-low-bit quantization constrains deployment on edge devices. Thus, the field still lacks unified strategies that jointly optimize communication efficiency, robustness, and privacy.

\textbf{RQ3 - Key Research Gaps and Open Issues:}  
The literature reveals several open gaps that must be addressed to realize the full potential of quantized FL. There is no standardized evaluation protocol for comparing quantization schemes, leading to inconsistent reporting of communication savings and convergence metrics~\cite{Gholami2021, caldas2018leaf}. Theoretical analyses of convergence behavior under quantization and non-IID data remain incomplete~\cite{Alistarh2017uniform, li2020federated}. Few studies incorporate real-world energy or hardware profiling, limiting practical relevance~\cite{nvidia2021tensorrt, pytorch2023quantization}.  Additionally, adaptive and personalized quantization—where clients employ device-specific or data-aware bit-widths—has received limited exploration. Addressing these gaps will require unified benchmarking, theoretical frameworks, and hardware–software co-design.

\textbf{RQ4 - Promising Future Directions:}  
Emerging trends point toward adaptive, hybrid, and hardware-aware quantization frameworks that balance accuracy, communication cost, and energy efficiency~\cite{Dong2019HAWQ, Wang2019HAQ}. Integrating quantization with pruning, sparsification, and differential privacy will enable holistic optimization of FL systems~\cite{Han2016, Bonawitz2017}. Furthermore, quantum-inspired and neuromorphic quantization paradigms offer novel avenues for ultra-efficient and secure distributed learning~\cite{Chris2020, Nguyen2025}. Future frameworks should incorporate multi-objective optimization to dynamically tune precision based on communication bandwidth, energy constraints, and privacy requirements.

\paragraph{Comparative Insights and Methodological Observations:} A key observation from the systematic review is the divergence between theoretical advances and practical implementations. While many quantized FL algorithms demonstrate strong empirical results in simulation, real-world validations on heterogeneous hardware are rare. Moreover, studies often focus narrowly on accuracy and bandwidth metrics, overlooking energy and latency trade-offs. In particular, adaptive quantization methods that adjust bit-widths per client or communication round have shown potential in reducing total energy consumption by up to 40\%~\cite{Dubey2024uniform}, yet lack consistent reproducibility due to dataset and system differences. Another insight concerns the complementary nature of PTQ and QAT. While PTQ is more suitable for low-power or resource-limited clients, QAT remains the preferred choice for applications demanding high reliability. Hybrid PTQ-QAT frameworks, which combine static quantization for stable layers with quantization-aware optimization for sensitive ones, represent a promising compromise between deployment simplicity and robustness.

\paragraph{Implications for Real-World FL Systems:} The findings highlight that quantization is not a universal solution but rather a configurable component within FL pipelines. For deployment at scale, quantization schemes must align with application-specific requirements, device capabilities, and privacy constraints. In edge intelligence scenarios—such as autonomous vehicles or medical IoT—quantization must maintain interpretability and fairness across clients. Hardware-aware quantization that considers the computational graph of accelerators (e.g., NPUs or TPUs) will be essential for efficient FL implementation~\cite{nvidia2021tensorrt, Gholami2021}. Moreover, integrating quantization with secure aggregation and differential privacy can ensure confidentiality while maintaining communication efficiency.

\paragraph{Limitations of This Review:} While this systematic survey follows the PRISMA methodology for transparency, several limitations remain. The scope of reviewed works is restricted to peer-reviewed publications available up to early 2025, potentially excluding recent preprints.  Finally, while this review emphasizes quantization, other model compression techniques—such as knowledge distillation and low-rank factorization—were not deeply explored. Future reviews could examine the synergy between quantization and broader compression paradigms for federated optimization.

Overall, quantization remains a cornerstone technique for achieving communication-efficient and scalable FL. While significant advancements have been made in quantization algorithms and federated adaptations, challenges persist in maintaining accuracy under heterogeneity, achieving theoretical convergence guarantees, and deploying models on constrained hardware. The next phase of research must emphasize adaptive precision control, energy-aware optimization, and rigorous benchmarking to transform quantized FL from a theoretical concept into a practical enabler for large-scale, privacy-preserving distributed intelligence.

\section{Conclusion}
\label{sec:conclusion}
This paper presented a comprehensive PRISMA-based review of quantization techniques in FL, analyzing their impact on communication efficiency, accuracy, and scalability. By organizing existing methods into PTQ and QAT paradigms, the study provides a unified framework for understanding quantization’s role in communication-efficient FL.
Regarding \textbf{RQ1}, findings show that quantization can achieve up to 10$\times$–50$\times$ compression with minimal accuracy loss when properly calibrated. PTQ ensures deployment simplicity, while QAT enhances robustness in heterogeneous and non-IID environments. Hybrid PTQ–QAT approaches offer promising trade-offs between efficiency and precision. For \textbf{RQ2}, major challenges include balancing communication efficiency and model accuracy, handling client heterogeneity, mitigating convergence instability, and ensuring privacy in quantized updates. Addressing these requires adaptive, bias-compensated, and privacy-preserving quantization techniques.
Under \textbf{RQ3}, research gaps persist in standardized evaluation, theoretical understanding of convergence, and real-world validation on edge hardware. The limited study of multi-objective, personalized, and privacy-preserving quantization frameworks highlights the field’s immaturity. Finally, \textbf{RQ4} points to future directions such as adaptive and hybrid quantization, integration with pruning and compression, energy- and hardware-aware methods, and quantum-inspired paradigms. Establishing open benchmarks and multi-objective optimization frameworks will be vital for scalable, sustainable FL.

In summary, quantization is a key enabler of efficient, privacy-preserving FL. Progress now depends on bridging theory and practice through adaptive precision, standardized evaluation, and hardware–software co-design to realize robust and energy-efficient distributed intelligence.



\newpage
\appendix
\section{Appendix A}

{
\footnotesize
\setlength{\tabcolsep}{4pt}
\renewcommand{\arraystretch}{1.15}
\raggedright

\begin{longtable}{
p{0.10\textwidth}  
p{0.15\textwidth}  
p{0.08\textwidth}  
p{0.12\textwidth}  
p{0.10\textwidth}  
p{0.10\textwidth}  
p{0.18\textwidth}  
}
\caption{Taxonomy of Quantization Techniques in FL based on Design Dimensions}
\label{tab:taxonomyFL_Quantization}\\
\textbf{Ref.} &
\textbf{Technique / Focus} &
\textbf{Training Stage} &
\textbf{Quantized Object} &
\textbf{Encoding Structure} &
\textbf{Precision Strategy} &
\textbf{Trade-off/ Application} \\

\midrule
\endfirsthead

\toprule
\textbf{Ref.} &
\textbf{Technique / Focus} &
\textbf{Training Stage} &
\textbf{Quantized Object} &
\textbf{Encoding Structure} &
\textbf{Precision Strategy} &
\textbf{Trade-off/Application} \\
\midrule
\endhead

\midrule
\multicolumn{7}{r}{\textit{Continued on next page}} \\
\endfoot

\bottomrule
\endlastfoot

AQFL~\cite{Abdelmoniem2021} &
Adaptive Bit &
PTQ &
Gradients &
Scalar &
Adaptive &
Accuracy-fairness trade-off; 
Mobile \& IoT devices \\

HAWQ~\cite{Danaee2022Biasquantization} &
Bias Correction &
QAT &
Model parameters &
Scalar &
Mixed &
Accuracy vs.\ energy, memory, latency;
Edge AI inference \\

GWEP~\cite{Prakash2022} &
Pruning+Quantization &
PTQ &
Weight + Gradient&
Scalar quantization (deterministic ternary / fixed-bit for weights, stochastic scalar quantization for gradients) &
Fixed-bit (typically INT8 / ternary) &
Reduce model redundancy and computation complexity;
Resource-constrained IoT device friendly \\

GFL~\cite{Thakur2024} &
Energy-Efficient Quantized FL &
PTQ &
Model parameters &
Scalar &
Adaptive &
Energy efficiency;
Green edge FL \\

HeteroSAg~\cite{Elkordy2022} &
Secure Heterogeneous Quantization &
PTQ &
Model updates &
Scalar &
Adaptive &
Privacy vs.\ aggregation complexity;
Privacy-preserving FL \\

FedPAQ\cite{Reisizadeh2020} &
Periodic / Uniform Quantization &
PTQ &
Model updates &
Scalar &
Fixed / Adaptive &
Lower uplink cost vs. delayed convergence;
Bandwidth-limited FL \\

FedSQ~\cite{Li2022_a} &
Stochastic Quantization &
PTQ &
Gradients &
Scalar &
Fixed-bit &
Unbiased updates vs. variance; convergence analysis; Communication efficiency; Wireless edge FL \\

QAT-FL.~\cite{Fang2026} &
Adaptive non-uniform quantization &
QAT &
Model updates (parameter differences) &
Vector (uniform / Lloyd–Max non-uniform) &
Adaptive-bit &
Lower quantization distortion vs. higher training complexity; Communication-efficient edge FL\\

HSQ~\cite{Dai2019} &
Hyper-Sphere Quantization &
PTQ &
Gradient &
Vector (codebook-based)&
Configurable / mixed (scalar pseudo-norm + discrete direction) &
communication efficiency vs.\ gradient accuracy; \\

SQFL~\cite{Marnissi2024}&
Adaptive Sparsification and Quantization &
PTQ &
Model parameters (local) + Model updates (uplink) & 
Scalar & 
Adaptive precision & 
Energy-efficient wireless FL\\

UVQ~\cite{Chen2021communication} &
Lattice / Vector Quantization &
PTQ &
Model updates &
Vector &
Fixed-bit &
Compression vs.\ aggregation noise;
Wireless FL \\

JoPEQ~\cite{Lang2023} &
Lattice + LDP &
PTQ &
Gradients &
Vector/Lattice &
Fixed-bit &
Privacy vs.\ utility;
Privacy-sensitive FL \\

DAdaQuant~\cite{Honig2022} &
Time- and Client-Adaptive &
PTQ &
Gradients &
Scalar &
Adaptive &
Efficiency vs. tuning complexity; non-IID FL \\

NQFL~\cite{Chen2024nqfl} &
Non-uniform Quantization (Lloyd–Max) &
PTQ &
Gradients &
Scalar &
Non-uniform &
Distribution-aware compression; IoT FL \\

LM-DFL~\cite{chen2024communication} &
Lloyd–Max adaptive vector quantization for decentralized FL &
PTQ &
Model updates (differential parameters) &
Vector (Lloyd–Max vector quantizer) &
Adaptive precision (doubly-adaptive quantization levels) &
Communication efficiency vs.\ quantization distortion and convergence accuracy; decentralized wireless edge federated learning \\

FedWSQ~\cite{Kim2025} &
Client-wise Non-uniform Quantization &
PTQ &
Model Parameters &
Scalar &
Non-uniform &
Accuracy vs. calibration cost; heterogeneous FL \\

GQFedWAvg~\cite{Li2024GQFedWAvg} &
Fairness-Aware Quantization &
PTQ &
Gradients &
Scalar &
Non-uniform &
Fairness vs. efficiency; edge FL \\

FedQNN~\cite{Ji2022} &
Sparse + Low-bit &
PTQ &
Model parameters &
Scalar &
Fixed-bit &
Compute and comm.\ efficiency;
IoT FL \\

FedVQCS~\cite{Oh2024} &
Vector Quant.\ + CS &
PTQ &
Gradients &
Vector &
Fixed-bit &
accuracy vs.\ bandwidth; 
Massive IoT FL \\

FedQCSO~\cite{Oh2022} &
Quantized Compressed Sensing &
PTQ &
Model updates &
Vector &
Fixed-bit &
Communication efficiency; 
Wireless FL \\

FedHieQ~\cite{Feng2022} &
Hierarchical Quantization &
PTQ &
Model updates &
Scalar &
Fixed-bit &
Efficiency vs.\ delay; 
Hierarchical FL \\

GCQ~\cite{Lin2021} &
Clipping + Quantization &
PTQ &
Gradients &
Scalar &
Fixed-bit &
Bias vs.\ convergence; 
Fading-channel FL \\

T-FedAvg~\cite{Xu2020} &
Ternary Quantization &
PTQ &
Gradients &
Scalar &
Fixed-bit &
Extreme compression vs. accuracy; 
Edge FL \\

FEDHBAA~\cite{Chen2023_(b)} &
Adaptive Bit Allocation &
PTQ &
Model Parameters &
Scalar &
Adaptive &
Convergence vs. accuracy; IoT FL \\

FedDQ~\cite{Qu2022} &
Descending Quantization &
PTQ &
Gradients &
Scalar &
Adaptive &
Communication savings vs. noise \\

LAQ~\cite{Sun2020} &
Lazy Aggregated Quantization &
PTQ &
Gradients &
Scalar &
Adaptive &
Communication vs.\ delay;
Wireless FL \\

AdaGQ~\cite{Liu2023_a} &
Gradient-Norm Adaptive Quantization &
PTQ &
Gradients &
Scalar &
Adaptive &
Stability vs. oscillations \\

QNN-FL~\cite{Kim2022} &
Fixed-Point Quantized NN &
PTQ &
Model Parameters &
Scalar &
Fixed-point &
Energy efficiency; embedded FL \\

Grace-FL~\cite{Thakur2025_a}&
Energy-aware adaptive gradient quantization with energy-weighted aggregation &
PTQ &
Gradient &
Scalar &
Adaptive &
Communication vs.\ Accuracy, Energy Efficiency vs.\ Convergence Speed, Energy Efficiency vs.\ Convergence Speed;
Green, resource-constrained federated learning \\

FedMPQ~\cite{chen2024mixed} &
Layer-wise Mixed Precision &
PTQ &
Model Parameters &
Scalar &
Mixed &
Efficiency vs. convergence \\

DP-FedQ~\cite{Gupta2022} &
Differentially Private Quantization &
PTQ &
Gradients &
Scalar &
Fixed-bit &
Privacy vs.\ accuracy; Private FL inference \\

Bozorgasl and Chen \cite{bozorgasl2025communication} &
Clipped uniform quantization with optimal clipping and stochastic quantization &
QAT &
Model parameters (weights) &
Scalar (uniform quantization with clipping) &
Fixed-bit precision &
Communication efficiency vs. quantization error; bandwidth-efficient federated learning for edge and IoT systems \\

ADQ~\cite{Chen2023_c} &
Adaptive Stochastic Quantization &
PTQ &
Gradients &
Scalar &
Adaptive &
Efficiency vs.\ convergence stability;
Bandwidth-limited FL \\

FedAQT~\cite{Ni2024} &
Energy-Constrained Quantization &
PTQ &
Gradients &
Scalar &
Fixed-bit &
Energy efficiency;
Mobile edge FL \\

EHFedQ~\cite{Ni2024_a} &
Energy-Harvesting FL &
PTQ &
Model parameters &
Scalar &
Adaptive &
Energy vs.\ convergence;
Wireless EH-FL \\

AdaQuantFL~\cite{Jhunjhunwala2021} &
Adaptive Uniform Quantization &
PTQ &
Gradients &
Scalar &
Adaptive &
Efficiency vs.\ stability;
Bandwidth-limited FL \\

OQFL~\cite{Kim2020} &
Learnable QAT &
QAT &
Model parameters &
Scalar &
Fixed-bit &
Accuracy vs.\ overhead; 
CNN-based FL \\

AGQFL~\cite{Lian2021} &
Sparse + Quantized Gradients &
PTQ &
Gradients &
Scalar &
Fixed-bit &
Efficiency vs.\ stability; 
Heterogeneous FL \\

RC-FED~\cite{Hamidi2025} &
Universal Quantization &
PTQ &
Gradients &
Scalar &
Fixed-bit &
Accuracy vs. \ Communication Cost;
Wireless Systems \\

Bouzinis \etal~\cite{Bouzinis2023} &
Stochastic Gradient Quant. &
PTQ &
Gradients &
Vector &
Adaptive &
Accuracy vs.\ Computation time, Convergence Analysis; Wireless n/w \\

LFL~\cite{amiri2020federatedlearningquantizedglobal} &
Quantized Global Updates &
PTQ &
Global parameters &
Scalar &
Fixed-bit &
Downlink efficiency; 
Large-scale FL \\

FEDDO~\cite{li2022joint} &
Joint Quantization \& Scheduling &
PTQ &
Model updates &
Scalar &
Adaptive &
Efficiency vs.\ optimization cost; 
Vehicular FL \\

FLQ~\cite{Tonellotto2021} & 
Random Quantization &
PTQ &
Model parameters &
Scalar &
Fixed-bit (uniform, binary, ternary, multi-bit) &
Accuracy vs.\ communication cost; Edge-cloud FL for IoT time-series\\

Shlezinger \etal~\cite{Shlezinger2020} &
Universal lattice-based vector quantization &
PTQ &
Model updates &
Vector lattice encoding &
Fixed-bit (rate-constrained) &
Communication vs. Accuracy;
Communication constrained FL\\

UVeQFed~\cite{Shlezinger2020_a} &
Universal Vector Quantization &
PTQ &
Model updates (weight differences)   &
Vector (lattice-based)     &
Fixed-bit precision (rate-constrained)     &
Communication vs. Accuracy; Bandwidth-limited FL     \\

Yu \etal~\cite{Yu2024} &
Block-wise QAT for Low-Bit Weights and Activations  &
QAT &
Weights + Activations &
Scalar (Uniform; Non-uniform compatible) &
Adaptive Sparsification and Quantization &
Low-bit CNN inference on resource-constrained devices \\

Ovi \etal~\cite{Ovi2022} &
Mixed-precision gradient quantization for privacy protection &
PTQ &
Gradient &
Scalar & 
Mixed Precision (Layer-wise) &
Privacy vs.\ Accuracy, Security vs.\ Complexity, Communication vs. Robustness; Resource- and bandwidth-constrained FL systems\\

FedQVR~\cite{Wang2025} &
Communication-efficient FL &
PTQ &
Model updates &
Vector (uniform / Lloyd–Max non-uniform) &
Fixed-bit (rate-constrained) &
Low computational complexity and easy deployment, but limited compression efficiency; preferred for resource-constrained edge and mobile FL.\\

DynFed~\cite{He2025} &
Dynamic bit-width quantization + knowledge distillation &
PTQ &
Gradients / model updates &
Scalar &
Adaptive &
Accuracy vs.\ communication efficiency under heterogeneous bit-widths; resource-aware federated learning for heterogeneous edge devices.\\

ARDıÇ \etal\cite{Emre2025} &
Adaptive Quantization + Differential Privacy &
PTQ &
Model updates / parameters &
Scalar (stochastic uniform) &
Adaptive &
Privacy + communication efficiency vs.\ accuracy degradation from quantization and DP noise; privacy-sensitive edge and medical FL \\

PP-QADMM~\cite{Elgabli2025} &
Dual-variable perturbation + stochastic quantized ADMM &
PTQ &
Model updates (primal + dual combination) &
Scalar (stochastic) &
Fixed-bit &
Privacy + communication efficiency vs.\ algorithmic complexity; wireless and privacy-sensitive FL \\

FedDT~\cite{He2025feddt} &
Knowledge distillation + ternary compression &
PTQ &
Model parameters (student model weights) &
Scalar (ternary encoding) &
Fixed-bit (ternary) &
Communication efficiency and heterogeneity mitigation vs.\ accuracy loss from aggressive compression; mobile and edge FL \\

BiPruneFL~\cite{Lee2025} &
Binary quantization + pruning-based FL &
PTQ &
Model parameters (binary weights) + gradients &
Scalar &
Binary (1-bit) &
Communication and computation efficiency vs.\ potential accuracy loss from aggressive compression; IoT and edge FL \\

HUFE-FL~\cite{Morais2025} &
Quantization + Huffman entropy coding &
PTQ &
Model parameters (weights + activations) &
Scalar + entropy encoding &
Fixed-bit (e.g., INT8) &
Communication reduction vs.\ encoding overhead; bandwidth-limited personalized FL and edge IoT systems \\

Fed\_IQ~\cite{Qin2025} &
Incremental model quantization with channel-aware uploading &
PTQ &
Gradients / model updates &
Scalar stochastic quantization &
Adaptive incremental precision &
Communication efficiency vs.\ quantization complexity; wireless FL in mobile edge and vehicular networks \\

FedCSTQ~\cite{Zheng2025} &
Compressed sensing + ternary quantization with heuristic gradient sparsification &
PTQ &
Gradients / model updates &
Vector (compressed sensing projection) &
Fixed-bit ternary precision &
Communication efficiency vs.\ reconstruction complexity; IoT and wireless FL \\

FedEnsemble~\cite{Ayman2026} &
Federated transformer ensemble with entropy-based attention stacking and QAT &
QAT &
Weights + activations &
Scalar &
Fixed-bit (4-bit / 8-bit) &
Communication efficiency vs.\ minor accuracy loss; privacy-preserving federated NLP and edge sentiment analysis \\

CEPAM~\cite{Shiu2026} &
Randomized vector quantization with differential privacy (CEPAM) &
PTQ &
Gradients &
Vector / lattice-based (LRSUQ) &
Fixed-bit stochastic (Gaussian/Laplace noise) &
Communication efficiency and privacy vs.\ added quantization noise; privacy-preserving FL in bandwidth-limited networks \\

MSPDQ-FL~\cite{Wang2026} &
Model-splitting privacy-preserving FL with dynamic quantization &
PTQ &
Model parameters (visible submodels) &
Scalar stochastic quantization &
Adaptive / dynamic precision &
Communication compression and privacy vs.\ protocol complexity; privacy-preserving FL in bandwidth-limited networks \\

PTQAT~\cite{Wang2025PTQAT} &
Hybrid PTQ + selective QAT fine-tuning &
PTQ + QAT &
Model parameters (weights) &
Scalar (uniform symmetric) &
Fixed low-bit (e.g., 4-bit weights) &
Accuracy close to QAT vs.\ lower training cost; efficient deployment of 3D perception models in autonomous driving and edge AI \\

\end{longtable}
}

In FL, model parameters denote the absolute network weights, model updates refer to the weight differences transmitted from clients to the server, and gradients correspond to local loss derivatives; although closely related, these quantities have distinct communication, convergence, and quantization properties and should be treated separately in FL system design.

\end{document}